\documentclass[preprint]{elsarticle}

\usepackage{hyperref}

\journal{}

\usepackage[process=auto]{pstool}
\usepackage{tikz}

\usepackage{stmaryrd}
\usepackage{amsfonts}

\usepackage[portuges,english]{babel}
\usepackage[utf8]{inputenc}

\usepackage{pdflscape}
\usepackage{graphicx,times,amsmath,amssymb} 
\usepackage{subfigure}
\usepackage[all]{xy}

\newcommand{\R}{\mathbb{R}}

\newcommand{\LL}{\mathbb{L}}

\newcommand{\vetx}{\mathbf{x}}
\newcommand{\vety}{\mathbf{y}}
\newcommand{\vetz}{\mathbf{z}}

\newcommand{\veta}{\mathbf{a}}
\newcommand{\vetb}{\mathbf{b}}

\newcommand{\NMSE}{\mbox{NMSE}}

\newcommand{\bb}{\begin{equation}}
\newcommand{\ee}{\end{equation}}
\newcommand{\bbb}{\begin{eqnarray}}
\newcommand{\eee}{\end{eqnarray}}
\newcommand{\benu}{\begin{enumerate}}
\newcommand{\eenu}{\end{enumerate}}

\newcommand{\vetw}{{\bf w}}

\newcommand{\bpm}{\begin{bmatrix}}
\newcommand{\epm}{\end{bmatrix}}

\def\boxmax{\kern 0em\hbox{\rm \kern .25em\lower.1ex\hbox{\rlap{$\vee$}}\kern -.07em\lower.2ex\hbox{$\square$}\kern.25em}}
\def\boxmin{\kern 0em\hbox{\rm \kern .25em\lower.1ex\hbox{\rlap{$\wedge$}}\kern -.07em\lower.2ex\hbox{$\square$}\kern.25em}}
\def\boxdiamond{\kern 0em\hbox{\rm \kern .25em\hbox{\rlap{$\diamond$}}\kern -.15em\lower.2ex\hbox{$\square$}}}

\newtheorem{thm}{Theorem}

\newtheorem{prop}{Proposition}
\newtheorem{exmp}{Example}
\newtheorem{rmrk}{Remark}
\newproof{pot}{Proof of Theorem \ref{theorem3}}
\newproof{pot1}{Proof of Theorem \ref{theorem4}}
\newproof{pot2}{Proof of Theorem \ref{theorem5}}

\graphicspath{{Figures/}}

\begin{document}

\begin{frontmatter}

\title{Max-$C$ and Min-$D$ Projection Autoassociative Fuzzy Morphological Memories: Theory and an Application for Face Recognition}



\author{Alex Santana dos Santos\corref{cor1}}
\ead{assantos@ufrb.edu.br}
\address{Exact and Technological Science Center, Federal University of Rec\^oncavo da Bahia, Rua Rui Barbosa, 710, Centro, Cruz das Almas-BA CEP 44380-000, Brazil}

\author{Marcos Eduardo Valle\corref{cor2}}
\ead{valle@ime.unicamp.br}
\address{Department of Applied Mathematics, University of Campinas, Rua Sérgio Buarque de Holanda, 651, Campinas-SP CEP 13083-859, Brazil}
\cortext[cor2]{Corresponding author}


%
%


\begin{abstract}
Max-$C$ and min-$D$ projection autoassociative fuzzy morphological memories (max-$C$ and min-$D$ PAFMMs) are two layer feedforward fuzzy morphological neural networks able to implement an associative memory designed for the storage and retrieval of finite fuzzy sets or vectors on a hypercube. In this paper we address the main features of these autoassociative memories, which include unlimited absolute storage capacity, fast retrieval of stored items, few spurious memories, and an excellent tolerance to either dilative noise or erosive noise. Particular attention is given to the so-called PAFMM of Zadeh which, besides performing no floating-point operations, exhibit the largest noise tolerance among max-$C$ and min-$D$ PAFMMs. Computational experiments reveal that Zadeh's max-$C$ PFAMM, combined with a noise masking strategy, yields a fast and robust classifier with strong potential for face recognition.
\end{abstract}

\begin{keyword}
Fuzzy associative memory\sep morphological neural network\sep lattice computing \sep face recognition.
\end{keyword}

\end{frontmatter}


\section{Introduction}
An associative memory (AM) is an input-output system inspired by the human brain ability to store and recall information by association \cite{hassoun97,kohonen89}. 
Apart from a large storage capacity, an ideal AM model should exhibit a certain tolerance to noise. In other words, we expect to retrieve a stored item not only presenting the original stimulus but also from a similar input stimulus \cite{hassoun97}. We speak of an autoassociative memory if stimulus and response coincide. For instance, our memory acts as an autoassociative model when we recognize a friend wearing sunglasses or a scarf. In other words, we obtain the desired output (recognize a friend) from a partial or noise input (its occluded face). We say that an associative memory is a heteroassociative model if at least one stimulus differs from its corresponding response. 
 
Several associative memory models have been introduced in the literature and their applications ranges from optimization \cite{hopfield85, serpen08} and prediction \cite{valle11nn,sussner18ins} to image processing and analysis \cite{valle11nn,grana16,sussner06fs,sussner17msc,valle15} and pattern classification \cite{esmi12hais,esmi15fs,esmi16fss,sussner06nn,valle16gcsr} including face recognition \cite{zhang04}. 

An AM model designed for the storage and recall of fuzzy sets on a finite universe of discourse is called a fuzzy associative memory (FAM) \cite{kosko92}. Since fuzzy sets can be interpreted as elements from a complete lattice \cite{goguen67} and mathematical morphology can be viewed as a theory on mappings between complete lattices \cite{heijmans94}, many important FAM models from the literature belong to the broad class of fuzzy morphological associative memories (FMAMs) \cite{valle11nn,valle08fss}. Briefly, FMAMs are implemented by fuzzy morphological neural networks. A morphological neural network is equipped with neurons that perform an elementary operation from mathematical morphology possibly followed by a non-linear activation function \cite{sussner11ins}. The class of FMAMs includes, for example, the max-mininum and max-product FAMs of Kosko \cite{kosko92}, the max-min FAM of Junbo et al. \cite{junbo94}, the max-min FAM with threshold of Liu \cite{liu99}, the fuzzy logical bidirectional associative memories of Belohlavek \cite{belohlavek00a}, and the implicative fuzzy associative memories (IFAMs) of Sussner and Valle \cite{sussner06fs}. In this paper, we only consider the max-$C$ and min-$D$ autoassociative fuzzy morphological memories (AFMMs) synthesized using the fuzzy learning by adjunction (FLA) \cite{valle11nn,valle08fss}. These autoassociative memories can be seen as fuzzy versions of the well-known matrix-based autoassociative morphological memories (AMMs) of \cite{ritter98}.




The main features of the max-$C$ and min-$D$ AFMMs are unlimited absolute storage capacity, one-step convergence when employed with feedback, and an excellent tolerance to either erosive or dilative noise. On the downside, the matrix-based AFMMs with FLA have a large number of spurious memories \cite{valle11nn}. A spurious memory is an item that has been unintentionally stored in the memory \cite{hassoun97}. Furthermore, the information stored on an AFMM with FLA is distributed on a synaptic weight matrix. As a consequence, these autossociative fuzzy memories consume a large amount of computational resources when designed for the storage and recall of large items \cite{valle11nn,vajg15}. 

Many autoassociative fuzzy memory models have been proposed in the literature to improve the noise tolerance or to reduce the computational cost of AFMMs with FLA. In order to improve the tolerance with respect to mixed noise, Valle developed the permutation-based finite IFAMs ($\pi$-IFAMs) by replacing the unit interval $[0,1]$ by a finite chain \cite{valle10ins}. A certain $\pi$-IFAM outperformed the original IFAMs in the reconstruction of gray-scale images corrupted by mixed salt and pepper noise. Similarly, to increase the noise tolerance of the IFAMs, Bui et al. introduced the so-called content-association associative memory (ACAM) \cite{bui15}. Using a fuzzy preorder relation, Perfilieva and Vajgl proposed a novel theoretical justification for IFAMs \cite{perfilieva15IFSA}. They also introduced a fast algorithm for data retrieval that is based on an IFAM model with a binary fuzzy preorder \cite{perfilieva16}. Moreover, Vajgl reduced the computational cost of an IFAM by replacing its synaptic weight matrix by a sparse matrix \cite{vajgl17}. In a similar fashion, the class of sparsely connected autoassociative fuzzy implications (SCAFIMs) is obtained by removing (possibly a significant amount of) synaptic weights from the original IFAMs \cite{valle10nc}. More generally, the quantale-based associative memories (QAMs) generalize several lattice-based autoassociative memories and have been effectively applied for the storage and the recall of large color images \cite{valle13prl}. Using piecewise linear transformation in the input and output spaces, Li et al. increased the storage capacity of fuzzy associative memories \cite{li17asc}. Recently, Sussner and Schuster proposed the interval-valued fuzzy morphological associative memories (IV-FMAMs) which are designed for the storage and retrieval of interval-valued fuzzy sets \cite{sussner18ins}. The novel IV-FMAMs have been effectively applied for time-series prediction.

Apart from the distributed models like the FMAMs with FLA and their variations, non-distributed associative memories models have received considerable attention in recent years partially due to their low computational effort and extraordinary successes in pattern recognition and image restoration tasks. Examples of non-distributed associative memories include models based on Hamming distance \cite{ikeda01} and kernels \cite{zhang04,souza18nafips} as well as subsethood and similarity measures  \cite{valle15,esmi15fs,esmi16fss,valle16gcsr,souza18tema}. In the context of non-distributed models, we recently introduced the max-plus and min-plus projection autoassociative morphological memories (max-plus and min-plus PAMMs) which can be viewed as non-distributed versions of the autoassociative morphological memories of Ritter et al. \cite{santos18nn,valle14wcciB}. Max-plus and min-plus PAMMs have less spurious memories than their corresponding distributed models and, thus, they are more robust to either dilative noise or erosive noise than the original autoassociative morphological memories. Computational experiments revealed that PAMMs and their compositions are competitive to other methods from the literature on classification tasks \cite{santos18nn}.

In the light of the successful developed of the max-plus and min-plus PAMMs and, in order to circumvent the aforementioned downsides of AFMMs, we introduced the class of max-$C$ projection autoassociative fuzzy morphological memories (max-$C$ PAFMMs) in the conference paper \cite{santos16cbsf}. Max-$C$ PAFMMs have been further discussed in \cite{santos17msc}, where some results concerning their implementation and storage capacity are given without proofs. In few words, a max-$C$ PAFMM projects the input into the family of all max-$C$ combinations of the stored items. In this paper, we present the dual version of max-$C$ PAFMMs: The class of min-$D$ PAFMMs which projects the input into the set of all min-$D$ combinations of the stored items \cite{santos17cnmac}. Furthermore, we address some theoretical results concerning both max-$C$ and min-$D$ PAFMM models. Thus, the theoretical part of this paper can be viewed as an extended version of the conference paper \cite{santos17cnmac}. In particular, we conclude that max-$C$ and min-$D$ PAFMMs exhibit better tolerance with respect to either dilative noise or erosive noise than their corresponding matrix-based AFMM with FLA. 
Additionally, we show in this paper that the most robust max-$C$ PAFMM with respect to dilative noise is based on Zadeh's inclusion measure and, thus, it is referred to as Zadeh's max-$C$ PAFMM. Accordingly, the dual of Zadeh's max-C PAFMM is the min-$D$ PAFMM most robust with respect to erosive noise. Finally, inspired by the work of Urcid and Ritter \cite{urcid07LC}, the frail tolerance of the max-$C$ and min-$D$ PAFMMs with respect to mixed noise can be improved significantly by masking the noise contained in the input \cite{santos17bracis}. Despite some preliminary experiments can be found in the conference paper \cite{santos17bracis}, we provide in this paper conclusive computational experiments concerning the application of Zadeh's max-$C$ PAFMM for face recognition.

The paper is organized as follows. Some basic concepts on fuzzy logic and fuzzy sets are briefly presented in next section. Section \ref{sec:AFMMs} briefly reviews the max-$C$ and min-$D$ AFMMs with FLA. The max-C and min-$D$ PAFMMs are addressed subsequently in Section \ref{sec:PAFMMs}. Zadeh's PAFMMs and the noise masking strategy are discussed in Sections \ref{sec:ZadehPAFMMs} and \ref{sec:noisemasking}, respectively. The performance of Zadeh's max-$C$ PAFMM for face recognition is addressed on Section \ref{sec:Experimentos}. We finish the paper with some concluding remarks and an appendix with the proofs of theorems.

\section{Some Basic Concepts on Fuzzy Systems}

The autoassociative fuzzy memories considered in this paper are based on fuzzy set theory and operations from fuzzy logic. In this section, we  briefly review the most important concepts on fuzzy systems. The interested reader is invited to consult \cite{barros17livro,klir95,nguyen00,gomide07,debaets97a} for a detailed review on this topic.

The key concept for the development of fuzzy morphological associative memories is adjunction \cite{heijmans95,deng02}. We say that  a fuzzy implication $I:[0,1] \times [0,1] \to [0,1]$ and a fuzzy conjunction $C:[0,1]\times [0,1] \to [0,1]$ form an adjunction  if the following equivalence holds true for $a,x,y \in [0,1]$:
\bb I(a,x) \geq y \Longleftrightarrow x \geq C(y,a).\ee
Analogously, a fuzzy disjunction $D:[0,1]\times [0,1]\to [0,1]$ and a fuzzy co-implication $J:[0,1]\times[0,1] \to [0,1]$ form an adjunction if and only if
\bb J(a,x) \leq y \Longleftrightarrow x \leq D(y,a).\ee
Examples of adjunction include the following pairs:
\begin{itemize}
 \item G\"odel's implication $I_M$ and the minimum fuzzy conjunction $C_M(x,y) = x \wedge y$.
 \item Goguen's implication $I_P$ and the product fuzzy conjunction $C_P(x,y) = x y$.
 \item Lukasiewicz's fuzzy implication $I_L(x,y) = 1 \wedge(1-x+y)$ and fuzzy conjunction $C_L(x,y)=0 \vee(x+y-1)$.
 \item Gaines' fuzzy implication and fuzzy conjunction defined respectively by
 \bb \label{eq:IGaines} I_{G}(x,y) = \begin{cases} 1, & x \leq y,\\ 0, & x>y, \end{cases} \quad \mbox{and} \quad  C_{G}(x,y)=\begin{cases} 0, & x =0,\\ y, & \mbox{otherwise}. \end{cases} \ee
 \item The maximum fuzzy disjunction $D_M(x,y) = x \vee y$ and G\"odel's fuzzy co-implication 
 \bb J_M(x,y) = \begin{cases} 0, & x \geq y,\\ y, & x<y. \end{cases} \ee
 \item The probabilistic sum disjunction $D(x,y) = x + y - xy$ and Goguen's fuzzy co-implication
 \bb J_P(x,y) =\begin{cases} 0, & x \geq y,\\ \dfrac{y-x}{1-y} , & x<y. \end{cases} \ee
 \item Lukasiewicz's disjunction $D_L(x,y) = 1 \wedge (x+y)$ and co-implication $J_L(x,y) = 0 \vee (y-x)$.
 \item Gaines' fuzzy disjunction and fuzzy co-implication defined as follows
 \bb \label{eq:JGaines}D_{G}(x,y)=\begin{cases} 1, & x =1,\\ y, & \mbox{otherwise} \end{cases} \quad \mbox{and} \quad  J_{G}(x,y) =\begin{cases} 0, & x \geq y,\\ 1, & x<y, \end{cases} \ee
\end{itemize}

We would like to point out that adjunctions arise naturally on complete lattices and are closely related to Galois connection and residuation theory \cite{birkhoff93,davey02,blyth72}. Furthermore, adjunction is one of the most important concept in mathematical morphology, a theory widely used for image processing and analysis \cite{heijmans95}. The elementary operations from mathematical morphology are erosions and dilations. Dilations and erosions are operators that commute with the supremum and infimum operations, respectively  \cite{heijmans94}. Formally, $\delta$ and $\varepsilon$ are respectively a dilation and an erosion if 
\bb \delta \left(\bigvee X \right) = \bigvee_{x \in X} \delta(x) \quad \mbox{and} \quad \varepsilon \left(\bigwedge X \right) = \bigwedge_{x \in X} \varepsilon(x),\ee
where the symbols ``$\bigvee$'' and ``$\bigwedge$'' denote respectively the supremum and infimum. 
It turns out that, if the pair $(I,C)$ forms an adjunction, then $I(a,\cdot)$ is an erosion and $C(\cdot,a)$ is a dilation for all $a \in [0,1]$. Moreover, if $C(\cdot,a)$ is a dilation for all $a \in [0,1]$, then there exists an unique fuzzy implication $I$ such that $(I,C)$ forms an adjunction. Such unique fuzzy implication that forms an adjunction with the fuzzy conjunction $C$ is the residual implication (R-implication) given by
\bb I(x,y) = \bigvee\{ t \in [0,1]: C(t,x) \leq y \}, \quad \forall x,y \in [0,1].\ee
Similarly, if the pair $(D,J)$ forms an adjunction then $D(\cdot,a)$ is an erosion and $J(a,\cdot)$ is a dilation. 
Also, if $D(\cdot,y)$ is an erosion for all $y \in [0,1]$, then its residual co-implication 
\bb J(x,y) =  \bigwedge\{ t \in [0,1]: D(t,x) \geq y\}, \quad \forall x,y \in [0,1]\},  \ee
is the unique fuzzy co-implication such that the pair $(D,J)$ forms an adjunction.

Apart from adjunctions, fuzzy logical operators can be connected by means of a strong fuzzy negation. A strong fuzzy negation is a nonincreasing mapping $\eta:[0,1] \to [0,1]$ such that $\eta(0)=1$, $\eta(1)=0$, and $\eta\big(\eta(x)\big)=x$ for all $x \in [0,1]$.
The standard fuzzy negation $\eta_S(x)=1-x$ is a strong fuzzy negation.
A fuzzy conjunction $C$ can be connected to a fuzzy disjunction $D$ by means of a strong fuzzy negation $\eta$ as follows:
\bb D(x,y)=\eta\Big(C\big(\eta(x),\eta(y)\big)\Big) \quad \mbox{or, equivalently,} \quad C(x,y)=\eta\Big(D\big(\eta(x),\eta(y)\big)\Big).\label{CdualD}\ee
In this case, we say that $C$ and $D$ are dual operators with respect to $\eta$. In a similar manner, a fuzzy co-implication $J$ is the dual operator of a fuzzy implication $I$
with respect to a strong fuzzy negation $\eta$ if and only if
\bb J(x,y)=\eta\Big(I\big(\eta(x),\eta(y)\big)\Big) \quad \mbox{or, equivalently,} \quad I(x,y)=\eta\Big(J\big(\eta(x),\eta(y)\big)\Big).\label{IdualJ}\ee
The pairs of fuzzy conjunction and fuzzy disjunction $(D_G, C_G)$, $(D_M, C_M)$, $(D_P, C_P)$, and $(D_L, C_L)$ are duals with respect the standard fuzzy negation $\eta_S$. The pairs $(I_G,J_G)$, $(I_M,J_M)$, $(I_P,J_P)$, and $(I_L,J_L)$ of fuzzy implication and fuzzy co-implication are also dual with relation the standard fuzzy negation.

The fuzzy logic operators $C$, $D$, $J$, and $I$ can be combined with either the maximum or the minimum operations to yield four matrix products. For instance, the max-$C$ and the min-$D$ matrix product of $A \in [0,1]^{m\times k}$ by $B \in [0,1]^{k\times n}$, denoted respectively by $G=A\circ B$ and $H=A \bullet B$, are defined by the following equations for all $i=1,\ldots,m$ and $j=1,\ldots,n$:
\bb  \label{prodmaxC} g_{ij}=\bigvee_{\xi=1}^k C(a_{i\xi},b_{\xi j}) \quad \mbox{and} \quad  h_{ij}=\bigwedge_{\xi=1}^k D(a_{i\xi},b_{\xi j}).\ee

In analogy to the concept of linear combination,  we say that $\mathbf{z} \in  [0,1]^n$ is a max-$C$ combination of the vectors belonging to the finite set $\mathcal{A}=\left\{\veta^1, \ldots, \veta^k\right\} \subset [0,1]^n$ if 
\bb \label{combinacao} \vetz = \bigvee_{\xi=1}^k C(\lambda_\xi,\veta^\xi) \ \Longleftrightarrow \ z_i = \bigvee_{\xi=1}^k C(\lambda_\xi,a_i^\xi), \forall i=1,\ldots,n,\ee
 where $\lambda_\xi \in [0,1]$ for all $\xi=1,\ldots,k$. Similarly, a min-$D$ combination of the vectors of  $\mathcal{A}$  is given by  \bb \label{combinacaoMin} \vety = \bigwedge_{\xi=1}^k D(\theta_\xi,\veta^\xi) \ \Longleftrightarrow \ y_i = \bigwedge_{\xi=1}^k D(\theta_\xi,a_i^\xi), \forall i=1,\ldots,n, \ee
where $\theta_\xi \in [0,1]$, for all $\xi=1,\ldots,k$. The sets of all max-$C$ combinations and min-$D$ combinations of $\mathcal{A}=\left\{\veta^1, \ldots, \veta^k\right\} \subset [0,1]^n$ are denoted respectively by 
\bb \label{eq:MaxCSet}\mathcal{C}(\mathcal{A})=\left\{\vetz=\bigvee_{\xi=1}^{k}C(\lambda_{\xi},\veta^{\xi}): \lambda_{\xi}\in [0,1]\right\}, \ee 
{and}
\bb\label{eq:MinDSet}\mathcal{D}(\mathcal{A})=\left\{\vetz=\bigwedge_{\xi=1}^{k}D(\theta_{\xi},\veta^{\xi}): \theta_{\xi}\in [0,1]\right\}.\ee 
The sets of max-$C$ and min-$D$ combinations plays a major role for the projection autoassociative fuzzy morphological memories (PAFMMs) presented in Section \ref{sec:PAFMMs}. Before, however, let us review the class of fuzzy autoassociative morphological memories which are defined using fuzzy logical connectives and adjunctions.

\section{Autoassociative Fuzzy Morphological Memories } \label{sec:AFMMs}

Let us briefly review the autoassociative fuzzy morphological memories (AFMM). The reader interested on a detailed account on this subject is invited to consult  \cite{valle11nn,valle08fss}.


Let $(I,C)$ and $(D,J)$ be adjunction pairs where $C$ is a fuzzy conjunction and $D$ is a fuzzy disjunction. As far as we know, most AFMMs are implemented by a single-layer network defined in terms of either the max-$C$ or the min-$D$ matrix products established by  \eqref{prodmaxC} \cite{valle11nn}. Formally, a max-$C$ and a min-$D$ autoassociative fuzzy morphological memory (AFMM) are mappings $\mathcal{W},\mathcal{M}: [0,1]^n \to [0,1]^n$ defined respectively by the following equations
 \bb \label{OutAFMM} \mathcal{W}(\vetx)=W\circ\vetx \quad \mbox{and} \quad \mathcal{M}(\vetx)=M\bullet\vetx, \quad \forall \vetx \in [0,1]^n, \ee 
where $W, M\in [0,1]^{n\times n}$ are called the synaptic weight matrices. The AFMMs $\mathcal{W}$ and $\mathcal{M}$ given by \eqref{OutAFMM} are called morphological because they perform respectively a dilation and an erosion from mathematical morphology.
Examples of AFMMs include the autoassociative version of the max-mininum and max-product fuzzy associative memories of Kosko \cite{kosko92}, the max–min fuzzy associative memories with threshold \cite{liu99}, and the implicative fuzzy associative memories \cite{sussner06fs}.

Let us now turn our attention to a recording recipe, called fuzzy learning by adjunction (FLA), which can be effectively used for the storage of  vectors on the AFMMs $\mathcal{W}$ and $\mathcal{M}$ defined by \eqref{OutAFMM} \cite{valle08fss}. Given a $\mathcal{A}=\left\{\veta^1,\ldots,\veta^k\right\} \subset [0,1]^n$, called the fundamental memory set, FLA determines the matrix $W  \in [0,1]^{n \times n}$ of a max-$C$ AFMM and the matrix $M \in [0,1]^{n \times n}$ of the min-$D$ AFMM by means of the following equations for all $i,j = 1,\ldots,n$:
\bb \label{eq:FLA} w_{ij} = \bigwedge_{\xi=1}^k I(a_j^\xi,a_i^\xi) \quad \mbox{and} \quad m_{ij} = \bigwedge_{\xi=1}^k J(a_j^\xi,a_i^\xi). \ee
We would like to point out that, although the matrices $M$ and $W$ given by \eqref{eq:FLA} are well defined for any fuzzy implication $I$ and fuzzy co-implication $J$, the following properties hold true only if they form an adjunction with the fuzzy conjunction $C$ and the fuzzy disjunction $D$, respectively.

We would like to point out that, using a strong fuzzy negation $\eta$, we can derive from a max-$C$ AFMM $\mathcal{W}$ another AFMM called the negation of $\mathcal{W}$ and denoted by $\mathcal{W}^*$. Formally, the negation  of $\mathcal{W}$ is defined by the equation
\bb \label{eq:negation} \mathcal{W}^*(\vetx) = \eta\Big( \mathcal{W}\big(\eta(\vetx)\big)\Big), \quad \forall \vetx \in [0,1]^n,\ee
where the strong fuzzy negation $\eta$ is applied in a component-wise manner. It is not hard to show that the negation of a max-$C$ AFMM $\mathcal{W}$ is a min-$D$ AFMM $\mathcal{M}$, and vice-versa, where the fuzzy conjunction $C$ and the fuzzy disjunction $D$ are dual with respect to the strong fuzzy negation $\eta$, i.e. they satisfy \eqref{CdualD} \cite{valle08fss}.


%

The following proposition reveals that a min-$D$ AFMM $\mathcal{M}$ and a max-$C$ AFMM $\mathcal{W}$, both synthesized using FLA, project the input $\vetx$ into the set of their fixed points \cite{valle11nn}. Furthermore, Proposition \ref{prop:fixed} shows that the output $\mathcal{M}(\vetx)$ of a min-$D$ AFMM with FLA is the largest fixed point less than or equal to the input $\vetx$. Analogously, a max-$C$ AFMM with FLA yields the smallest fixed point which is greater than or equal to the input \cite{valle11nn}.

\begin{prop}[Valle and Sussner \cite{valle11nn}] \label{prop:fixed} Let $(I,C)$ and $(D,J)$ be adjunction pairs where $C$ and $D$ are respectively an associative fuzzy conjunction and an associative fuzzy disjunction, both with a left identity. In this case, the output of the min-$D$ AFMM $\mathcal{M}$ defined by \eqref{OutAFMM}  with FLA given by \eqref{eq:FLA}  satisfies 
\bb \label{ProjFAMM} \mathcal{M}(\vetx)=\bigvee\left\{\vetz \in \mathcal{I}(\mathcal{A}): \vetz\leq\vetx\right\}, \quad \forall \vetx \in [0,1]^n, \ee
where $\mathcal{I}(\mathcal{A})$ denotes the set of all fixed points of $\mathcal{M}$ which depends on and includes the fundamental memory set $\mathcal{A} = \{\veta^1,\ldots,\veta^k\}$. Dually, the output of the max-$C$ AFMM $\mathcal{W}$ with FLA satisfies 
\bb \label{ProjFAMMw} \mathcal{W}(\vetx)=\bigwedge\left\{\vety \in \mathcal{J}(\mathcal{A}): \vety\geq\vetx\right\}, \quad \forall \vetx \in [0,1]^n, \ee 
where $\mathcal{J}(\mathcal{A})$ denotes the set of all fixed points of $\mathcal{W}$ which also depends on and contains the fundamental memory set $\mathcal{A}$.
\end{prop}

In the light of Proposition \ref{prop:fixed}, besides the adjunction relationship, from now on we assume that the fuzzy disjunction $D$ and the fuzzy conjunction $C$ are associative and have both a left identity. As a consequence, AFMMs with FLA present the following properties: they can store as many vectors as desired; they have a large number of spurious memories; an AFMM exhibits tolerance to either dilative noise or erosive noise, but it is extremely sensitive to mixed (dilative+erosive) noise. Recall that a distorted version $\vetx$ of a fundamental memory $\veta^{\xi}$ has undergone a dilative change if $\vetx\geq\veta^{\xi}$. Dually, we say that $\vetx$ has undergone an erosive change if $\vetx\leq\veta^{\xi}$ \cite{ritter98}.

\begin{exmp} \label{Example1}
Consider the fundamental memory set
\bb \label{eq:ex_memoria} \mathcal{A}= \left\{
\veta^1 = \left[\begin{array}{c}    0.4\\   0.3\\    0.7\\0.2 \end{array}\right], 
\veta^2 = \left[\begin{array}{c}  0.1\\  0.7\\    0.5\\ 0.8  \end{array}\right],
\veta^3 = \left[\begin{array}{c}   0.8\\    0.5\\    0.4\\    0.2\\ \end{array}\right] \right\}.\ee
Using Gödel's co-implication $J_M$ in \eqref{eq:FLA}, the synaptic weight matrix $M_M$ of the min-$D_M$ AFMM $\mathcal{M}_M$ with FLA is:
\bb \label{MatrixM_M} M_M = \begin{bmatrix}
 0.00& 0.80 &0.80 & 0.80\\
  0.70 &  0.00  & 0.70  &0.50\\
0.70 &0.70&0.00& 0.70\\
 0.80 & 0.80& 0.80 & 0.00
\end{bmatrix}.\ee
Now, consider the input fuzzy set \bb \label{eq:ex_input} \vetx = \begin{bmatrix}0.4 \ & 0.3 \ & 0.8 \ & 0.7 \end{bmatrix}^T.\ee
Note that $\vetx$ is a dilated version of the fundamental memory $\veta^1$ because $\vetx = \veta^1 + [0.0 \ 0.0 \ 0.1 \ 0.5]^T \geq \veta^1$. 
The output of the min-$D_M$ AFMM with FLA is 
\bb\mathcal{M}_M(\vetx)=M_M\bullet_M\vetx= \begin{bmatrix} 0.40 & 0.30 & 0.70  &0.70 \end{bmatrix}^T\neq \veta^1,\label{exOutM}\ee
where ``$\bullet_M$'' denotes the min-$D_M$ product defined in terms of the fuzzy disjunction $D_M$. According to Proposition \ref{prop:fixed}, the output $\begin{bmatrix} 0.40 & 0.30 & 0.70  &0.70 \end{bmatrix}^T$ is a fixed point of $\mathcal{M}_M$ that does not belong to the fundamental memory set $\mathcal{A}$. Thus, it is a spurious memory of $\mathcal{M}_M$. In a similar fashion, we can use FLA to store the fundamental set $\mathcal{A}$ into the min-$D$ AFMMs  $\mathcal{M}_{P}$, $\mathcal{M}_{L}$, and $\mathcal{M}_{G}$ obtained by considering respectively the probabilistic sum, the Lukasiewicz, and the Gaines fuzzy disjunction. Upon presentation of the input vector $\vetx$ given by \eqref{eq:ex_input}, the min-$D$ AFMMs $\mathcal{M}_{P}$, $\mathcal{M}_{L}$, and $\mathcal{M}_{G}$ yield respectively
\bb\mathcal{M}_P(\vetx)=M_P\bullet_P\vetx= \begin{bmatrix} 0.40& 0.30& 0.70& 0.53 \end{bmatrix}^T\neq \veta^1,\label{exOutM_P}\ee
\bb\mathcal{M}_L(\vetx)=M_L\bullet_L\vetx= \begin{bmatrix} 0.40&   0.30&    0.70& 0.40 \end{bmatrix}^T\neq \veta^1,\label{exOutM_L}\ee
and
\bb \mathcal{M}_G(\vetx)=M_L\bullet_L\vetx= \begin{bmatrix} 0.40&0.30&0.80&0.70 \end{bmatrix}^T\neq \veta^1\label{exOutM_G}.\ee
Such as the min-$D_M$ AFMM $\mathcal{M}_M$, the autoassociative memories $\mathcal{M}_{P}$, $\mathcal{M}_{L}$, and $\mathcal{M}_{G}$ failed to produce the desired output $\veta^1$. 
\end{exmp} 

\section{Max-$C$ and Min-$D$ Fuzzy Projection Autoassociative Morphological Memories} \label{sec:PAFMMs}

As distributed $n \times n$ matrix-based autoassociative memories, a great deal of computer memory is consumed by min-$D$ and max-$C$ AFMMs if the length $n$ of stored vectors is large. Furthermore, from Proposition \ref{prop:fixed}, their tolerance with respect to either dilative or erosive noise is degraded as the number of fixed points increase. 

Inspired by the feature that min-$D$ and max-$C$ AFMMs with FLA project the input vector into the set of their fixed point, we can improve the noise tolerance of these memory models by reducing their set of fixed points. Accordingly, we recently introduced the max-$C$ projection autoassociative fuzzy memories (max-$C$ PAFMMs) by replacing in \eqref{ProjFAMM} the set $\mathcal{I}(\mathcal{A})$ by the set $\mathcal{C}(\mathcal{A})$ of all max-$C$ combinations  of vectors of $\mathcal{A}$ \cite{santos16cbsf,santos17msc}. Formally, given a set $\mathcal{A}=\left\{\veta^1,\ldots,\veta^k\right\} \subset [0,1]^n$, a max-$C$ PAFMM $\mathcal{V}:[0,1]^n \rightarrow [0,1]^n$ is defined by
\bb \label{ProjV} \mathcal{V}(\vetx)=\bigvee\left\{\vetz \in \mathcal{C}(\mathcal{A}) : \vetz \leq \vetx \right\}, \quad \forall \vetx \in [0,1]^n,  \ee  where the set $\mathcal{C}(\mathcal{A})$ is defined in \eqref{eq:MaxCSet}. 
A dual model, referred to as min-$D$ PAFMM, is obtained by replacing $\mathcal{J}(\mathcal{A})$ by the set $\mathcal{D}(\mathcal{A})$ of all min-$D$ combinations of the fundamental memories in \eqref{ProjFAMMw}. Precisely, a min-$D$ PAFMM $\mathcal{S}:[0,1]^n \rightarrow [0,1]^n$ satisfies
\bb \label{ProjS} \mathcal{S}(\vetx)=\bigwedge\left\{\vety \in \mathcal{D}(\mathcal{A}) : \vety \geq \vetx \right\},\quad \forall \vetx \in [0,1]^n, \ee where the set $\mathcal{D}(\mathcal{A})$ is given in \eqref{eq:MinDSet}. 
The following theorem is a straightforward consequence of these definitions.
\begin{thm}\label{theorem2}  The max-$C$ and min-$D$ PAFMMs given respectively by \eqref{ProjV} and \eqref{ProjS} satisfy the inequalities $\mathcal{V}(\vetx)\leq\vetx\leq\mathcal{S}(\vetx)$ for any input vector $\vetx \in [0,1]^n$. Furthermore, $\mathcal{V}(\mathcal{V}(\vetx))=\mathcal{V}(\vetx)$ and  $\mathcal{S}(\mathcal{S}(\vetx))=\mathcal{S}(\vetx)$ for all $\vetx \in [0,1]^n$.  
\end{thm}

As a consequence of Theorem \ref{theorem2}, a max-$C$ PAFMM and a min-$D$ PAFMM are respectively an opening and a closing form fuzzy mathematical morphology \cite{deng02}. Like the min-$D$ AFMM, a max-$C$ PAFMM exhibits only tolerance with respect to dilative noise. Also, it is extremely sensitive to either erosive or mixed noise. In fact, a fundamental memory $\veta^\xi$ cannot be retrieved by a max-$C$ PAFMM from an input such that $\vetx \leq \veta^\xi$. In a similar manner, such as the max-$C$ AFMM, a min-$D$ PAFMM $\mathcal{S}$ exhibits tolerance with respect to erosive noise but it is not robust to either dilative or mixed noise. 

Let us now address the absolute storage capacity of max-$C$ and min-$D$ PAFMMs. Clearly, a max-$C$ PAFMM has optimal absolute storage capacity if a fundamental memory $\veta^\xi$ belongs to $\mathcal{C}(\mathcal{A})$. In other words, if $\veta^\xi \in \mathcal{C}(\mathcal{A})$, then $\mathcal{V}(\veta^\xi)=\veta^\xi$. It turns out that $\veta^\xi$ belongs to the set of all max-$C$ combinations of $\veta^1,\ldots,\veta^k$ if the fuzzy conjunction $C$ has a left identity, i.e., there exists $e \in [0,1]$ such that $C(e,x)=x$ for all $x \in [0,1]$. In fact, for any fuzzy conjunction $C$, we have $C(0,x)=C(x,0)=0$ for all $x\in [0,1]$ \cite{debaets97a}. Thus, if the fuzzy conjunction $C$ has a left identity, we can express a fundamental memory $\veta^\xi$ by the following max-$C$ combination 
\bb\veta^\xi= C(0,\veta^1)\vee \ldots \vee C(e,\veta^\xi)\vee \ldots \vee C(0,\veta^k).\ee
Dually, a min-$D$ PAFMM has optimal absolute storage capacity if the fuzzy disjunction $D$ has a left identity. Summarizing, we have the following theorem:

\begin{thm} \label{theorem1} Let $C$ and $D$ denote respectively a fuzzy conjunction and a fuzzy disjunction and consider a fundamental memory set $\mathcal{A}=\left\{\veta^1,\ldots,\veta^k\right\} \subset [0,1]^n$.
The max-$C$ PAFMM given by \eqref{ProjV} satisfies $\mathcal{V}(\veta^\xi)=\veta^\xi$ for all  $\xi \in \mathcal{K}$ if the fuzzy conjunction $C$ has a left identity. Dually, if the fuzzy disjunction $D$ has a left identity then $\mathcal{S}(\veta^\xi)=\veta^\xi$ for all $\xi \in \mathcal{K}$, where $S$ denotes the min-$D$ PAFMM given by \eqref{ProjS}. 
\end{thm}

The next theorem, which is based on adjunctions, provides effective formulas for the implementation of the max-$C$ and min-$D$ PAFMMs. 

\begin{thm} \label{theorem3}
Given a fundamental memory set $\mathcal{A}=\left\{\veta^1, \ldots, \veta^k\right\} \subset [0,1]^{n}$.
Let a fuzzy implication $I$ and a fuzzy conjunction $C$ form an adjunction. For any input $\vetx \in [0,1]^n$, the max-$C$ PAFMM $\mathcal{V}$ given by \eqref{ProjV} satisfies
\bb\mathcal{V}(\vetx)= \bigvee_{\xi=1}^{k}C(\lambda_{\xi},\veta^{\xi}), \quad \mbox{where} \quad \lambda_{\xi}=\bigwedge_{j=1}^{n}I(a^{\xi}_j,x_j). \label{forPontual}\ee 
Dually, let a fuzzy disjunction $D$ and a fuzzy co-implication $J$ form an adjunction. For any input $\vetx \in [0,1]^n$, the output of the min-$D$ PAFMM $\mathcal{S}$ can be computed by
\bb\mathcal{S}(\vetx)= \bigwedge_{\xi=1}^{k}D(\theta_{\xi},\veta^{\xi}), \quad \mbox{where} \quad \theta_{\xi}=\bigvee_{j=1}^{n}J(a^{\xi}_j,x_j). \label{forPontualS}\ee  
\end{thm}

In the light of Proposition \ref{prop:fixed}, we only consider PAFMMs based fuzzy conjunctions and fuzzy disjunctions that form adjunction pairs with a fuzzy implication and a fuzzy co-implication, respectively.

\begin{rmrk} Theorem \ref{theorem3} above gives a formula for the coefficients $\lambda_\xi$ that is used to define the output of a max-$C$ PAFMM. Note that the coefficient $\lambda_{\xi}$ corresponds to the  degree of inclusion of the fundamental memory $\veta^{\xi}$ in the input fuzzy set $\vetx$. In other words, we have
\bb \label{eq:lambda_inc} \lambda_\xi = Inc_{\mathcal{F}}(\veta^\xi,\vetx), \quad \forall \xi \in \mathcal{K},\ee 
where $Inc_\mathcal{F}$ denotes the Bandler-Kohout fuzzy inclusion measure \cite{bandler80}.
\end{rmrk}

As to the computational effort, max-$C$ and min-$D$ PAFMMs are usually cheaper than their corresponding min-$D$ and max-$C$ AFMMs. In fact, from \eqref{theorem3}, max-$C$ and min-$D$ PAFMMs are non-distributive memory models which can be implemented by fuzzy morphological neural networks with a single hidden layer \cite{valle08fss,sussner08gr,valle18wiley}. They do not require the storage of a synaptic weight matrix of size $n \times n$. Also, they perform less floating-point operations than their corresponding min-$D$ and max-$C$ AFMMs if $k<n$. To illustrate this remark, consider a fundamental memory set $\mathcal{A}=\left\{\veta^1,\ldots,\veta^k\right\}$, where $\veta^{\xi} \in [0,1]^{n}$ for all $\xi \in \{1,\ldots,k\}$ with $k < n$. On the one hand, to synthesize the synaptic weight matrix of a min-$D$ AFMM $\mathcal{M}$, we perform $kn^2$ evaluation of a fuzzy co-implication and $(2k-1)n^2$ comparisons. Besides, the resulting synaptic weight matrix consumes $\mathcal{O}(n^2)$ of memory space. In the recall phase, the min-$D$ AFMM $\mathcal{M}$ requires $n^2$ evaluations of a fuzzy disjunction and $(2n-1)n$ comparisons. On the other hand, to compute the parameters $\lambda$'s of a max-$C$ PAFMM $\mathcal{V}$, we perform $2nk$ evaluations of a fuzzy implication and $(2n-1)k$ comparisons. The subsequent step of the max-$C$ PAFMM $\mathcal{V}$ requires $2nk$ evaluations of a fuzzy conjunction and $(k-1)n$ comparisons. Lastly, it consumes $\mathcal{O}(nk)$ of memory space for the storage of the fundamental memories. Similar remarks holds for a max-$C$ AFMM and a min-$D$ PAFMM. Concluding, Table \ref{tab:Floating} summarizes the computational effort in the recall phase of the AFMMs and PAFMMs. Here, fuzzy operations refers to evaluations of fuzzy conjunction, disjunction, implication, or co-implications.

\begin{table}
\begin{center}
\begin{tabular}{|l|l|l|l|} \hline 
 & Fuzzy Operations & Comparison & Memory Space \\ \hline
 AFMMs $\mathcal{M}$ and $\mathcal{W}$ & $\mathcal{O}(n^2)$&$\mathcal{O}(n^2)$ & $\mathcal{O}(n^2)$\\ \hline
PAFMMs $\mathcal{V}$ and $\mathcal{S}$ & $\mathcal{O}(nk)$&$\mathcal{O}(nk)$ &  $\mathcal{O}(nk)$\\ \hline
\end{tabular}
\caption{Computational complexity in the recall phase of autoassociative memories} \label{tab:Floating}
\end{center}
\end{table}

Finally, different from the min-$D$ and max-$C$ AFMMs, the max-$C$ and min-$D$ PAFMMs are not dual models with respect to a strong fuzzy negation. 
Precisely, the next theorem shows that the negation of a min-$D$ PAFMM is a max-$C$ PAFMM designed for the storage of the negation of the fundamental memories, and vice-versa.

\begin{thm}\label{theorem4}
Let $(I,C)$ and $(D,J)$ be adjunction pairs where the fuzzy conjunction $C$ is connected to the fuzzy disjunction $D$ by means of a strong fuzzy negation $\eta$, that is, $C$, $D$, and $\eta$ satisfies \eqref{CdualD}. Given a fundamental memory set $\mathcal{A}=\left\{\veta^1,\ldots,\veta^k\right\} \subset [0,1]^{n}$, define $\mathcal{B} = \{\mathbf{b}^1,\ldots,\mathbf{b}^k\}$ by setting $b_i^\xi = \eta(a_i^\xi)$, for all $i=1,\ldots,n$ and $\xi \in \mathcal{K}$. Also, let $\mathcal{V}$ and $\mathcal{S}$ be respectively the max-$C$ and the min-$D$ PAFMMs designed for the storage of $\veta^1,\ldots,\veta^k$ and define their negation as follows for every $\vetx \in [0,1]^n$:
 \bb \mathcal{V}^{*}(\vetx)=\eta(\mathcal{V}[\eta(\vetx) ]) \quad \mbox{and} \quad \mathcal{S}^{*}(\vetx)=\eta(\mathcal{S}[\eta(\vetx) ]) .\label{PminDdual}\ee
 The negation $\mathcal{S}^*$ of $\mathcal{S}$ is the max-$C$ PAFMM designed for the storage of $\mathbf{b}^1,\ldots,\mathbf{b}^k$, that is, \bb \label{eq:dual} \mathcal{S}^{*}(\vetx)=\bigvee_{\xi=1}^k C(\lambda_{\xi}^*, \mathbf{b}^{\xi}), \quad \mbox{where} \quad \lambda_{\xi}^{*}=\bigwedge_{j=1}^{n}I(b^{\xi}_j,x_j).\ee
Analogously, the negation $\mathcal{V}^*$ of $\mathcal{V}$ is the min-$D$ PAFMM given by \bb\mathcal{V}^{*}(\vetx)=\bigwedge_{\xi=1}^k D(\theta_{\xi}^*, \mathbf{b}^{\xi}), \quad \mbox{where} \quad \theta_{\xi}^{*}=\bigvee_{j=1}^{n}J(b^{\xi}_j,x_j).\ee
\end{thm}
It follows from Theorem \ref{theorem4} that the negations $\mathcal{S}^{*}$ and $\mathcal{V}^{*}$ fail to store the fundamental memory set $\mathcal{A}=\left\{\veta^1,\ldots,\veta^k\right\}$.

\begin{exmp} \label{ExGodel} Consider the fundamental memory set $\mathcal{A}$ given by \eqref{eq:ex_memoria}. Let $C_M$ and $I_M$ be the minimum fuzzy conjunction and  fuzzy implication of G{\"o}del, respectively. We synthesized the max-$C$ PAFMM $\mathcal{V}_M$ designed for the storage of $\mathcal{A}$ using the adjunction pair $(I_M, C_M)$. Since $C_M$ is a fuzzy conjunction with 1 as identity, from Theorem \ref{theorem1}, the equation $\mathcal{V}_M(\veta^\xi)=\veta^\xi$ holds for $\xi=1,2,3$. Given the input vector $\vetx$ defined by \eqref{eq:ex_input}, we obtain from \eqref{forPontual} the coefficients 
\bb \lambda_1 = 1.0, \quad \lambda_2 =  0.3, \quad \mbox{and} \quad \lambda_3= 0.3.\ee
Thus, the output of the max-$C$ PAFMM $\mathcal{V}_M$ is
\bbb \mathcal{V}_M(\vetx) &=& C_M(\lambda_1 , \veta^1) \vee C_M(\lambda_2,\veta^2)  \vee C_M(\lambda_3,\veta^3) \nonumber\\
&=& \begin{bmatrix}   0.40  &  0.30 & 0.70&  0.30\end{bmatrix}^T\neq \veta^1.\eee
Note that $\mathcal{V}_M$ failed to retrieve the fundamental memory $\veta^1$.

Analogously, we can store the fundamental memory set $\mathcal{A}$ into the max-$C$ PAFMMs $\mathcal{V}_{P}$ and $\mathcal{V}_{L}$ using respectively the adjunction pairs $(I_P, C_P)$ and $(I_L, C_L)$. Upon presentation of the vector $\vetx$, the max-$C$ PAFMMs $\mathcal{V}_{P}$ and $\mathcal{V}_{L}$ produce 
\bb \mathcal{V}_P(\vetx) =  \begin{bmatrix}  0.40& 0.30& 0.70& 0.34\end{bmatrix}^T\neq \veta^1\ee
and
\bb \mathcal{V}_L(\vetx) =  \begin{bmatrix}   0.40& 0.30&0.70& 0.40 \end{bmatrix}^T\neq \veta^1.\ee
Like the min-$D$ AFMMs $\mathcal{M}_P$ and $\mathcal{M}_L$, the memories $\mathcal{V}_{P}$ and $\mathcal{V}_{L}$ failed to recall the fundamental memory $\veta^1$. Nevertheless, the max-$C$ PAFMMs $\mathcal{V}_M$, $\mathcal{V}_{P}$, and $\mathcal{V}_{L}$ yielded outputs are more similar to the desired vector $\veta^1$ than the min-$D$ AFMMs $\mathcal{M}_M$, $\mathcal{M}_{P}$, $\mathcal{M}_{L}$, and $\mathcal{M}_{G}$ (see Example \ref{Example1}). Quantitatively, Table \ref{tab:NMSE} shows the normalized mean squared error (NMSE) between the recalled vector and the desired output $\veta^1$. Recall that the NMSE between $\vetx$ and $\veta$ is given by
\bb \label{eq:NMSE} \NMSE(\veta,\vetb)  = \frac{\|\vetx-\veta\|_2^2}{\|\veta\|_2^2} = \frac{\sum_{j=1}^n (x_j - a_j)^2}{\sum_{j=1}^n a_j^2}. \ee
This simple example confirms that a max-$C$ PAFMM can exhibit a better tolerance with respect to dilative noise than its corresponding min-$D$ AFMM. 

\begin{table}
\begin{center}
\scalebox{0.8}{\begin{tabular}{|c|l|l|l|l|l|l|l|l|l|}  \hline 
$ \bullet$&$\vetx$ &$\mathcal{M}_M(\vetx)$& $\mathcal{M}_P(\vetx)$&$\mathcal{M}_L(\vetx)$& $\mathcal{M}_G(\vetx)$&$\mathcal{V}_M(\vetx)$& $\mathcal{V}_P(\vetx)$&$\mathcal{V}_L(\vetx)$&$\mathcal{V}_\mathcal{Z}(\vetx)$\\ \hline
 $\NMSE(\bullet,\veta^1) $& 0.33&    0.32  &  0.14 &   0.05& 0.33  & 0.01   & 0.02 &   0.05& \textbf{0.00} \\ \hline
\end{tabular}}
\caption{Normalized mean squared error.} \label{tab:NMSE}
\end{center}
\end{table} 


\end{exmp}

Let us conclude the section by emphasizing that we cannot ensure optimal absolute storage capacity if $C$ does not have a left identity.

\begin{exmp} Consider the ``compensatory and'' fuzzy conjunction  defined by \bb C_A(x,y) = \sqrt{(xy)(x+y-xy)}. \ee  
Note that $C_A$ does not have a left identity. Moreover, the fuzzy implication that forms an adjunction with $C_A$ is 
\bb I_A(x,y) =\begin{cases} 1, & x =0,\\ 1\wedge \left[\dfrac{-x^2 +\sqrt{x^2+4x(1-x)y^2}}{2x(1-x)}\right], & 0<x<1, \\ y^2, & x =1. \end{cases} \ee
Now, let $\mathcal{V}_A:[0,1]^4 \to [0,1]^4$ be the max-$C_A$ PAFMM designed for the storage of the fundamental memory set $\mathcal{A}$ given by \eqref{eq:ex_memoria}. Upon the presentation of the fundamental memory $\veta^1$ as input, we obtain from \eqref{forPontual} the coefficients
\bb \lambda_1=0.39,\quad \lambda_2=0.06 \quad \mbox{and} \quad \lambda_3=0.23.\ee
Thus, the output vector of the max-$C$ PAFMM $\mathcal{V}_A$ is
\bb\mathcal{V}_A(\veta^1) = C_A(\lambda_1 , \veta^1) \vee C_A(\lambda_2,\veta^2)  \vee C_A(\lambda_3,\veta^3)=\begin{bmatrix} 0.40\\0.27\\0.47\\0.20 \end{bmatrix}\neq \veta^1.\ee
In a similar fashion, using the fundamental memories $\veta^2$ and $\veta^3$ as input, we obtain from $\mathcal{V}_A$ the outputs
\bb \mathcal{V}_A(\veta^2) =\begin{bmatrix} 0.10\\0.39 \\0.30\\ 0.44 \end{bmatrix}\neq\veta^2 \quad \mbox{and} \quad \mathcal{V}_A(\veta^3) =\begin{bmatrix} 0.52\\ 0.37\\ 0.40\\0.20 \end{bmatrix}\neq\veta^3. \ee
In accordance with Theorem \ref{theorem2}, the inequality $\mathcal{V}_A(\veta^{\xi})\leq\veta^{\xi}$ holds for $\xi=1,2,3$. The fundamental memories $\veta^1$, $\veta^2$, and $\veta^3$, however, are not fixed points of the max-$C$ PAFMM  $\mathcal{V}_A$.
\end{exmp}

\section{Zadeh Max-$C$ PAFMM and Its Dual Model}\label{sec:ZadehPAFMMs}

According to Hassoun and Watta \cite{hassoun97}, one of the most common problems in an associative memory design task is the creation of spurious or false memories. A spurious memory is a fixed point of an autoassociative memory that does not belong to the fundamental memory set. For instance, the fixed point $\vety = [0.4,0.3,0.7,0.3]^T$ is a spurious memory of the max-$C$ PAFMM $\mathcal{V}_M$ in Example \ref{ExGodel}. 

In general, the noise tolerance of an autoassociative memory decreases as the number of spurious memories increase. The set of fixed points of a max-$C$ PAFMM, however, corresponds to the set of all max-$C$ combinations of the fundamental memories. Hence, the smaller the family $\mathcal{C}(\mathcal{A})$, the higher the noise tolerance of a max-$C$ PAFMM. 

Given a fundamental memory set $\mathcal{A}=\{\veta^1,\ldots,\veta^k\}$, we can reduce $\mathcal{C}(\mathcal{A})$ significantly by considering in \eqref{eq:MaxCSet} the fuzzy conjunction of Gaines $C_G$. From Theorem \ref{theorem1}, the output of the max-$C$ PAFMM based on Gaines' fuzzy conjunction is given by 
\bb \label{eq:max-C_G} \mathcal{V}_\mathcal{Z}(\vetx)= \bigvee_{\xi=1}^{k}C_G(\lambda_{\xi},\veta^{\xi}),\ee
where
\bb \lambda_\xi = \bigwedge_{i=1}^n I_G(a^\xi_j,x_j) = Inc_\mathcal{Z}(\veta^\xi,\vetx), \ \forall \xi \in \mathcal{K},\ee
where $Inc_{\mathcal{Z}}:[0,1]^n \times [0,1]^n \to [0,1]$ is the fuzzy inclusion measure of Zadeh defined as follows for all $\veta,\vetb \in [0,1]^n$:
\bbb Inc_{\mathcal{Z}}(\veta,\vetb) &=&\begin{cases} 1, & a_j \leq b_j, \forall j=1,\ldots,n, \\ 0, & \mbox{otherwise}. \end{cases} \label{IncZadeh}\eee
In other words, instead of a general fuzzy inclusion measure of Bander-Kohout, the coefficients $\lambda_\xi$ are determined using Zadeh's fuzzy inclusion measure $Inc_\mathcal{Z}$. Hence, this max-$C$ PAFMM is referred to as Zadeh's max-$C$ PAFMM and denoted by $\mathcal{V}_\mathcal{Z}$. 

From \eqref{IncZadeh}, the coefficient $\lambda_\xi=Inc_\mathcal{Z}(\veta^\xi,\vetx)$ is either 0 or 1. Moreover, $\lambda_\xi=1$ if and only if $a_j^\xi \leq x_j$ for all $j=1,\ldots,n$. Also, we have $C_G(0,x)=0$ and $C_G(1,x)=x$ for all $x \in [0,1]$. Therefore, for any input $\vetx \in [0,1]^n$, the output of Zadeh's max-$C$ PAFMM is alternatively given by the equation
\bb \label{ProjVZadeh} \mathcal{V}_\mathcal{Z}(\vetx)= \bigvee_{\xi \in \mathcal{I}} \veta^\xi,\ee
where 
\bb \label{eq:setI} \mathcal{I}=\{\xi: a_j^\xi \leq x_j, \forall j=1,\ldots,n\},  \ee
is the set of the indexes $\xi$ such that $\veta^\xi$ is less than or equal to the input $\vetx$, i.e., $\veta^\xi \leq \vetx$. Here, we have $\mathcal{V}_\mathcal{Z}(\vetx) = \mathbf{0}$ if $\mathcal{I}=\emptyset$, where $\mathbf{0}$ is a vector of zeros.

In a similar manner, from \eqref{forPontualS}, the dual of Zadeh's max-$C$ PAFMM is the min-$D$ PAFMM defined by 
\bb\mathcal{S}_\mathcal{Z}(\vetx)= \bigwedge_{\xi=1}^{k}D_G(\theta_{\xi},\veta^{\xi}), \quad \mbox{where} \quad \theta_{\xi}=\bigvee_{j=1}^{n}J_G(a^{\xi}_j,x_j). \label{eq:min-D_G}\ee  
Here, $D_G$ and $I_G$ denote respectively the fuzzy disjunction and fuzzy co-implicantion of Gaines. Alternatively,  the output of Zadeh's min-$D$ PAFMM is given by 
\bb \label{ProjSZadeh} \mathcal{S}_\mathcal{Z}(\vetx)= \bigwedge_{\xi \in \mathcal{J}} \veta^\xi,\ee
where 
\bb \label{eq:setJ} \mathcal{J}=\{\xi: a_j^\xi \geq x_j, \forall j=1,\ldots,n\},  \ee
is the set of indexes $\xi$ such that $\veta^\xi \geq \vetx$. Here, we have $\mathcal{S}_\mathcal{Z}(\vetx) = \mathbf{1}$ if $\mathcal{J}=\emptyset$,  where $\mathbf{1}$ is a vector of ones.

Note from \eqref{ProjVZadeh} and \eqref{ProjSZadeh} that no arithmetic operation is performed during the recall phase of Zadeh's max-C PAFMM model and its dual model; they only perform comparisons! Thus,  both $\mathcal{V}_{\mathcal{Z}}$ and $\mathcal{S}_{\mathcal{Z}}$ are  computationally cheap and fast associative memories. In addition, Zadeh's max-$C$ PAFMM is extremely robust to dilative noise. On the other hand, its dual model $\mathcal{S}_{\mathcal{Z}}$ exhibits an excellent tolerance with respect to erosive noise. The following theorem address the noise tolerance of these memory models.

\begin{thm} \label{theorem5} Consider the fundamental memory set $\mathcal{A}= \{\veta^1,\ldots,\veta^k\} \subset [0,1]^n$. The identity $\mathcal{V}_{\mathcal{Z}}(\vetx)=\veta^{\gamma}$ holds true if there exists an unique $\gamma \in \mathcal{K}$ such that $\veta^{\gamma} \leq \vetx$. Furthermore, if there exists an unique $\mu \in \mathcal{K}$ such that $\veta^{\mu} \geq \vetx$ then $\mathcal{S}_{\mathcal{Z}}(\vetx)=\veta^{\mu}$.

 \end{thm}

\begin{exmp} \label{ExZadeh} Consider the fundamental memory set $\mathcal{A}$ given by \eqref{eq:ex_memoria} and the input fuzzy set $\vetx$ defined by \eqref{eq:ex_input}. Clearly, $\veta^1 \leq \vetx$, $\veta^2 \not\leq \vetx$, and  $\veta^3 \not\leq \vetx$. Thus, the set of indexes defined by \eqref{eq:setI} is $\mathcal{I}=\{1\}$. From \eqref{ProjVZadeh}, the output of Zadeh's max-$C$ PAFMM is 
\bb \mathcal{V}_\mathcal{Z}(\vetx) = \bigvee_{\xi \in \mathcal{I}} \veta^\xi = \veta^1.\ee
Note that the max-$C$ PAFMM $\mathcal{V}_\mathcal{Z}$ achieved perfect recall of the original fundamental memory. As a consequence, the NMSE is zero. From  Table \ref{tab:NMSE}, the max-$C$ PAFMM of Zadeh yielded the best NMSE, followed by the max-$C$ PAFMMs $\mathcal{V}_{M}$, $\mathcal{V}_{P}$ and $\mathcal{V}_{L}$. 
\end{exmp}

Let us conclude this section by remarking that Zadeh's max-$C$ PAFMM also belongs to the class of $\Theta$-fuzzy associative memories ($\Theta$-FAMs) proposed by Esmi et al. \cite{esmi15fs}. 

\begin{rmrk} 
An autoassociative $\Theta$-FAM is defined as follows: Consider a fundamental memory set $\mathcal{A}=\{\veta^1,\ldots,\veta^k\} \subset [0,1]^n$ and let $\Theta^\xi:[0,1]^n \to [0,1]$ be operators such that $\Theta^\xi(\veta^\xi)=1$ for all $\xi=1,\ldots,k$. Given an input $\vetx$ and a weight vector $\vetw = [w_1,\ldots,w_k] \in \R^k$, a $\Theta$-FAM $\mathcal{O}$ yields 
\[ \mathcal{O}(\vetx) = \bigvee_{\xi \in \mathcal{I}_\vetw(\vetx)} \veta^\xi,\]
where $\mathcal{I}_\vetw(\vetx)$ is the following set of indexes:
\[ \mathcal{I}_\vetw(\vetx) = \{\eta: w_\eta \Theta^\eta(\vetx) = \max_{\xi=1:k} w_\xi \Theta^\xi(\vetx)\}.\]
Now, the max-$C$ PAFMM of Zadeh is obtained by considering $\vetw=[1,1,\ldots,1] \in \R^k$ and $\Theta^\xi(\cdot) = Inc_\mathcal{Z}(\veta^\xi,\cdot)$, for all $\xi=1,\ldots,k$. Specifically, in this case $\mathcal{I}_\vetw(\vetx)$ coincides with the set of index $\mathcal{I}$ defined by \eqref{ProjVZadeh}.
\end{rmrk}

\section{Noise Masking Strategy for PAFMMs} \label{sec:noisemasking}

A max-$C$ PAFMM cannot retrieve a fundamental memory $\veta^\xi$ from an input $\vetx \leq \veta^\xi$. The frail tolerance with respect to erosive or mixed noise may limit the applications of a max-$C$ PAFMM to real world problems. From the duality principle, similar remarks holds true for min-$D$ PAFMMs. It turns out that the noise tolerance of a PAFMM can be significantly improved by masking the noise contained in a corrupted input \cite{urcid07LC}. In few words, noise masking converts an input degraded by mixed noise into a vector corrupted by either dilative or erosive noise. Inspired by the works of Urcid and Ritter \cite{urcid07LC}, let us present a noise masking strategy for the PAFMMs. In order to simplify the presentation, we shall focus on max-$C$ PAFMMs.

Let $\mathcal{V}$ denote a max-$C$ PAFMM which have been synthesized using a fundamental memory set $\mathcal{A}=\{\veta^1,\ldots,\veta^k\}$. Also, assume that $\vetx$ is a version of the fundamental memory $\veta^\eta$ corrupted by mixed noise. Then, $\veta_d^\eta = \vetx \vee \veta^\eta$ is the masked input vector which contains only dilative noise, i.e., the inequality $\veta_d^\eta \geq \veta^\eta$ holds true. Since $\mathcal{V}$ is robust to dilative noise, we expect the max-$C$ PAFMM to be able to retrieve the original fuzzy set $\veta^\eta$ under presentation of the masked vector $\veta_d^\eta$. 

The noise masking idea has a practical shortcoming: we do not known a priori which fundamental memory have been corrupted.
Hence, Urcid and Ritter suggested to compare, for all $\xi \in \mathcal{K}$, the masked vector $\veta_d^\xi = \vetx \vee \veta^\xi$ with both the input $\vetx$ and the fundamental memory $\veta^\xi$ \cite{urcid07LC}. The comparison is based on some meaningful measure such as the normalized mean squared error (NMSE).  In this paper, we propose to use a fuzzy similarity measure to determine the masked vector. Briefly, a fuzzy similarity measure is a mapping $\sigma:[0,1]^n \times [0,1]^n \to [0,1]$ such that $\sigma(\veta,\vetb)$ corresponds to the degree of similarity between $\veta \in [0,1]^n$ and $\vetb \in [0,1]^n$  \cite{couso13,debaets05,baets09,fan99a,xuecheng92}. Using a fuzzy similarity measure, the masked vector $\veta_d^\eta$ is obtained by computing the maximum between the input $\vetx$ and the fundamental memory $\veta^\eta$ most similar to the input. In mathematical terms, we have $\veta^\eta_d = \vetx \vee \veta^\eta$ where $\eta$ is an index such that
\bb \label{eq:minSM} \sigma(\vetx,\veta^\eta)  = \bigvee_{\xi=1}^k  \left\{\sigma(\vetx,\veta^\xi)\right\}.\ee

Concluding, the technique of noise masking for recall of vectors using a max-$C$ PAFMM $\mathcal{V}$ yields the autoassociative fuzzy morphological memory $\mathcal{V}^M:[0,1]^n \rightarrow [0,1]^n$ defined by \bb \label{eq:noisemasking} \mathcal{V}^M(\vetx)=\mathcal{V}(\vetx \vee \veta^\eta), \quad \forall \vetx \in [0,1]^n, \ee 
where $\eta$ is an index that satisfies \eqref{eq:minSM}. 

In a similar manner, we can define the technique of noise masking for recall of vectors using a min-$D$ PAFMM $\mathcal{S}$. Formally, we denote by 
$\mathcal{S}^M$ an autoassociative fuzzy morphological memory given by
 \bb \label{eq:noisemasking1} \mathcal{S}^M(\vetx)=\mathcal{S}(\vetx \wedge \veta^\eta),\ee 
where $\eta$ is an index that satisfies \eqref{eq:minSM} and $\mathcal{S}:[0,1]^n \to [0,1]^n$ is a min-$D$ PAFMM. 

In the previous section, we pointed out that the Zadeh max-$C$ PAFMM $\mathcal{V}_{\mathcal{Z}}$, as well as its dual model, does not perform floating point arithmetic operations. Some arithmetic operations, however, may be required for the computation of the masked input fuzzy set. For instance, if we consider in \eqref{eq:minSM} the Hamming similarity measure $\sigma_H$ defined by 
\bb \label{eq:SimilarityH} \sigma_H(\veta, \vetb)=1 -\dfrac{1}{n}\sum_{i=1}^{N}\left|a_i-b_i\right|,  \quad \forall \veta,\vetb \in [0,1]^n,\ee 
then the memory  $\mathcal{V}^M_{\mathcal{Z}}$ performs $(2n+1)k$ floating point operations during the retrieval phase.

\section{Computational Experiments} \label{sec:Experimentos}


Inspired by the autoassociative memory-based classifiers described in \cite{sussner06nn,zhang04}, we propose the following autoassociative memory-based classifier for face images. Suppose we have a training dataset with $k_i$ different face images from an individual $i$, for $i=1,\ldots,c$. Each face image is encoded into a column-vector $\veta^{\xi,i} \in [0,1]^n$, where $i \in \{1,\ldots,c\}$ and $\xi \in \{1,\ldots,k_i\}$. We shall address below two approaches to encode face images into $[0,1]^n$. For now, let $\mathcal{M}^i$ denote an autoassociative memory designed for the storage of the fundamental memory set $\mathcal{A}^i = \{\veta^{1,i},\ldots,\veta^{k_i,i}\} \subset [0,1]^n$ composed by all training images from individual $i \in \{1,\ldots, c\}$. Given an unknown face image, we also encode it into a column-vector $\vetx \in [0,1]^{n}$ using the same procedure as the training images. Then, we present $\vetx$ as input to the autoassociative memories $\mathcal{M}^i$'s. Finally, we assign the unknown face image to the first individual $\eta$ such that 
\bb \label{eq:classifier} \sigma \big(\vetx, \mathcal{M}^\eta(\vetx) \big) \geq \sigma \big( \vetx, \mathcal{M}^i(\vetx) \big), \quad \forall i=1,\ldots,c,\ee
where $\sigma$ denotes a fuzzy similarity measure. In words, $\vetx$ belongs to an individual such that the recalled vector is the most similar to the input. 

In our experiments, we used in \eqref{eq:classifier} the Hamming similarity measure defined by \eqref{eq:SimilarityH}. Furthermore, a face image have been encoded into a column-vector $\vetx \in [0,1]^n$ using the following two approaches: 
\begin{enumerate}
 \item First, by using the {\tt MATLAB}-style sequence of commands {\tt rgb2gray}\footnote{This command is applied only if the input is a color face image in the RGB color space.}, {\tt im2double}, {\tt imresize}, and {\tt reshape}. We would like to point out that we resized the images according to the dimensions used by Feng et al. \cite{feng17}.
 \item Secondly, using a variation of the ResNet proposed by He et al. \cite{He16} followed by a data transformation. Precisely,  we used the {\tt python} package {\tt face\_recognition} which is based on {\tt Dlib} library \cite{Dlib}. Briefly, the {\tt face\_recognition} package is equipped with a pre-trained ResNet deep network with 29 convolutional layers trained by deep residual learning \cite{He16}. The ResNet network maps a face image into a vector $\mathbf{v} \in \R^{128}$ such that $\|\mathbf{v}\|_2= 1$. We obtained $\vetx \in [0,1]^n$ by applying the following transformation where $\mu_i$ and $\sigma_i$ denote the mean and standard deviation of the $i$th component of all 128-dimensional training vectors.
 \bb x_i = \frac{1}{1+e^{-(v_i - \mu_i)/\sigma_i}}, \quad \forall i =1,\ldots, 128. \ee
\end{enumerate}


Apart from models from the literature, we only consider the min-$D_L$ AFMM and Zadeh's max-$C$ PAFMM. Recall that the AFMMs based on the Lukasiewicz connectives outperformed many other AFMMs on experiments concerning the retrieval of gray-scale images \cite{sussner06fs}. Furthermore, the min-$D_L$ AFMM can be obtained from the morphological autoassociative memory of Ritter et al. using thresholds \cite{sussner06fs,ritter98}. As pointed out in Section \ref{sec:ZadehPAFMMs}, Zadeh's PAFMM is expected to exhibit a tolerance with respect to dilative noise larger than the other max-$C$ PAFMMs. Concluding, we synthesized the classifiers based on the min-$D$ AFMM $\mathcal{M}_L^M$ and Zadeh's max-$C$ PAFMM $\mathcal{V}_\mathcal{Z}^M$, both equipped with the noise masking strategy described by \eqref{eq:noisemasking} and \eqref{eq:SimilarityH}. These classifiers, combined with the first encoding strategy listed above are denoted respectively by Resized+$\mathcal{M}_L^M$ and Resized+$\mathcal{V}_\mathcal{Z}^M$. Similarly, we refer to ResNet+$\mathcal{M}_L^M$ and ResNet+$\mathcal{V}_\mathcal{Z}^M$ the fuzzy associative memory-based classifiers combined with the second approach listed above. 

The performance of the $\mathcal{M}_L^M$- and $\mathcal{V}_\mathcal{Z}^M$-based classifiers have been compared with the following approaches from the literature: sparse representation classifier (SRC) \cite{wright09}, linear regression-based classifier (LRC) \cite{naseem10}, collaborative representation-based classifier (CRC) \cite{zhang11}, fast superimposed sparse parameter (FSSP(1)) classifier \cite{feng17}, and the deep network ResNet available at the {\tt python} {\tt face\_recognition} package. 

\subsection{Face Recognition with Expressions and/or Pose}

Face recognition has been an active research topic in pattern recognition and computer vision due to its applications in human computer interaction, security, access control, and others \cite{zhang04,zhang11,feng17}. To evaluate the performance of the $\mathcal{V}_\mathcal{Z}^M$-based classifier, we conducted experiments using three standard face databases. Namely, \textit{Georgia Tech Face Database} (GT) \cite{GTdatabase}, \textit{AT$\&$T Face Database} \cite{ORLdatabase}, and \textit{AR Face Image Database} \cite{martinez98a}. These face databases incorporate pose, illumination, and gesture alterations.

\begin{itemize}
\item The Georgia Tech (GT) Face Database contains face images of 50 individuals taken in two or three sessions at the Center for Signal and Image Processing at  Georgia Institute of Technology  \cite{GTdatabase}. These images, which sum up to 15 per individual, show frontal and/or tilted faces with different facial expressions, lighting conditions, and scale. In this paper, we used the cropped images available at the GT dataset. Figure \ref{fig:GT} presents the 15 facial images of one individual from the GT database. As pointed out previously, the color images from the cropped GT database have been converted into gray-scale face images and resized to $30 \times 40$ pixels before presented to the classifiers SRC, LRC, CRC, FSSP(1), Resized+$\mathcal{M}_L^M$, and Resized+$\mathcal{V}_\mathcal{Z}^M$. 

\begin{figure}[t]
\begin{center}
\begin{tabular}{cccccc}
\includegraphics[width=1.8cm, height = 2.4cm]{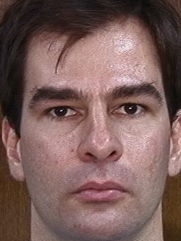} & \includegraphics[width=1.8cm, height = 2.4cm]{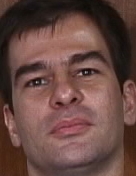}& \includegraphics[width=1.8cm, height = 2.4cm]{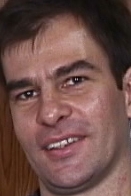}& \includegraphics[width=1.8cm, height = 2.4cm]{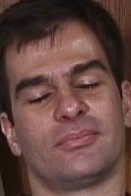}& \includegraphics[width=1.8cm, height = 2.4cm]{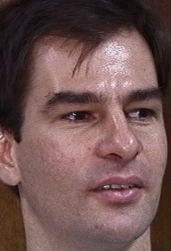}\\ 
\includegraphics[width=1.8cm, height = 2.4cm]{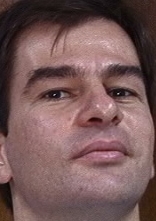} & \includegraphics[width=1.8cm, height = 2.4cm]{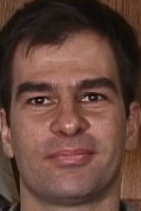}& \includegraphics[width=1.8cm, height = 2.4cm]{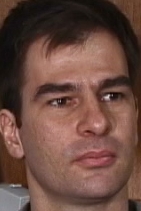}& \includegraphics[width=1.8cm, height = 2.4cm]{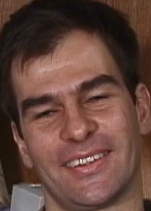}& \includegraphics[width=1.8cm, height = 2.4cm]{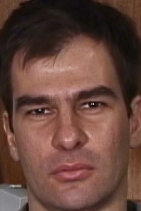}\\ 
\includegraphics[width=1.8cm, height = 2.4cm]{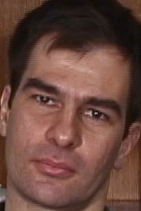} & \includegraphics[width=1.8cm, height = 2.4cm]{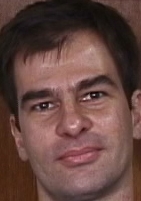}& \includegraphics[width=1.8cm, height = 2.4cm]{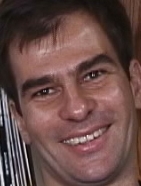}& \includegraphics[width=1.8cm, height = 2.4cm]{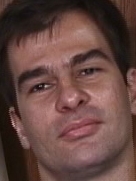}& \includegraphics[width=1.8cm, height = 2.4cm]{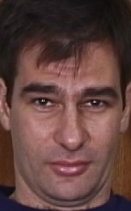}\\
\end{tabular}
\caption{Images of one individual from the GT face database}\label{fig:GT}
\end{center}
\end{figure}

\item The AT$\&$T database, formerly known as the \textit{ORL database of faces}, has 10 different images for each of 40 distinct individuals \cite{ORLdatabase}. All face images are in up-right and frontal position. The 10 images of an individual is shown in Figure \ref{fig:ORL} for illustrative purposes. For the classifiers SRC, LRC, CRC, FSSP(1), Resized+$\mathcal{M}_L^M$, and Resized+$\mathcal{V}_\mathcal{Z}^M$ , the face images of the AT$\&$T database have been resized to $28\times 23$ pixels. 
\begin{figure}[t]
\begin{center}
\begin{tabular}{cccccc}
\includegraphics[width=1.8cm, height = 2.2cm]{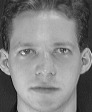} & \includegraphics[width=1.8cm, height = 2.2cm]{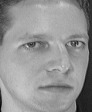}& \includegraphics[width=1.8cm, height = 2.2cm]{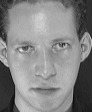}& \includegraphics[width=1.8cm, height = 2.2cm]{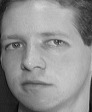}& \includegraphics[width=1.8cm, height = 2.2cm]{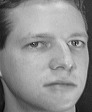}\\ 
\includegraphics[width=1.8cm, height = 2.2cm]{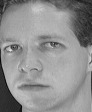} & \includegraphics[width=1.8cm, height = 2.2cm]{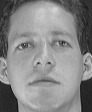}& \includegraphics[width=1.8cm, height = 2.2cm]{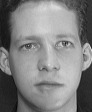}& \includegraphics[width=1.8cm, height = 2.2cm]{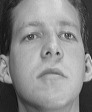}& \includegraphics[width=1.8cm, height = 2.2cm]{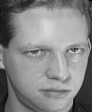}\\ 
\end{tabular}
\caption{Images of one individual from the AT$\&$T face database}\label{fig:ORL}
\end{center}
\end{figure}

\item The AR face image database contains over 4000 facial images from 126 individuals \cite{martinez98a}. For each individual, 26 images have been taken in two different sessions separated by two weeks. The face images features from different facial expressions, illumination changes, and occlusions. In our experiments, we used a subset of the cropped AR face image database with face images of 100 individuals. Furthermore, we only considered the 8 face images with different expressions (normal, smile, anger, and scream) from each individual. The 8 face images of one individual of the AR database is shown in Figure \ref{fig:AR}. Finally, we would like to point out that the images in AR database have been converted to gray-scale images and resized  to $50 \times 40$ pixels for the classifiers SRC, LRC, CRC, FSSP(1), Resized+$\mathcal{M}_L^M$, and Resized+$\mathcal{V}_\mathcal{Z}^M$.  

\begin{figure}[t] 
\begin{center}
\begin{tabular}{cccc} 
\includegraphics[width=1.8cm, height = 2.4cm]{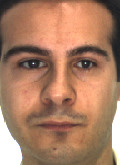} & \includegraphics[width=1.8cm, height = 2.4cm]{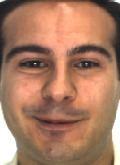}& \includegraphics[width=1.8cm, height = 2.4cm]{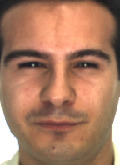}& \includegraphics[width=1.8cm, height = 2.4cm]{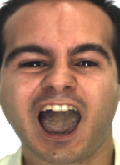}\\ \includegraphics[width=1.8cm, height = 2.4cm]{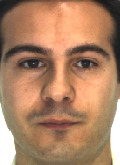}& \includegraphics[width=1.8cm, height = 2.4cm]{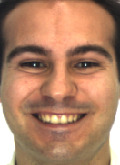}& \includegraphics[width=1.8cm, height = 2.4cm]{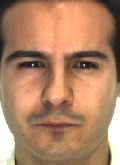}& \includegraphics[width=1.8cm, height = 2.4cm]{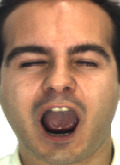}\\
\end{tabular}
\caption{Some images of one individual from the AR face database. }\label{fig:AR}
\end{center}
\end{figure} 
\end{itemize}

\begin{figure}
\begin{tabular}{c}
a) Box-plot of normalized recognition rates. \\
\includegraphics[width=0.9\columnwidth]{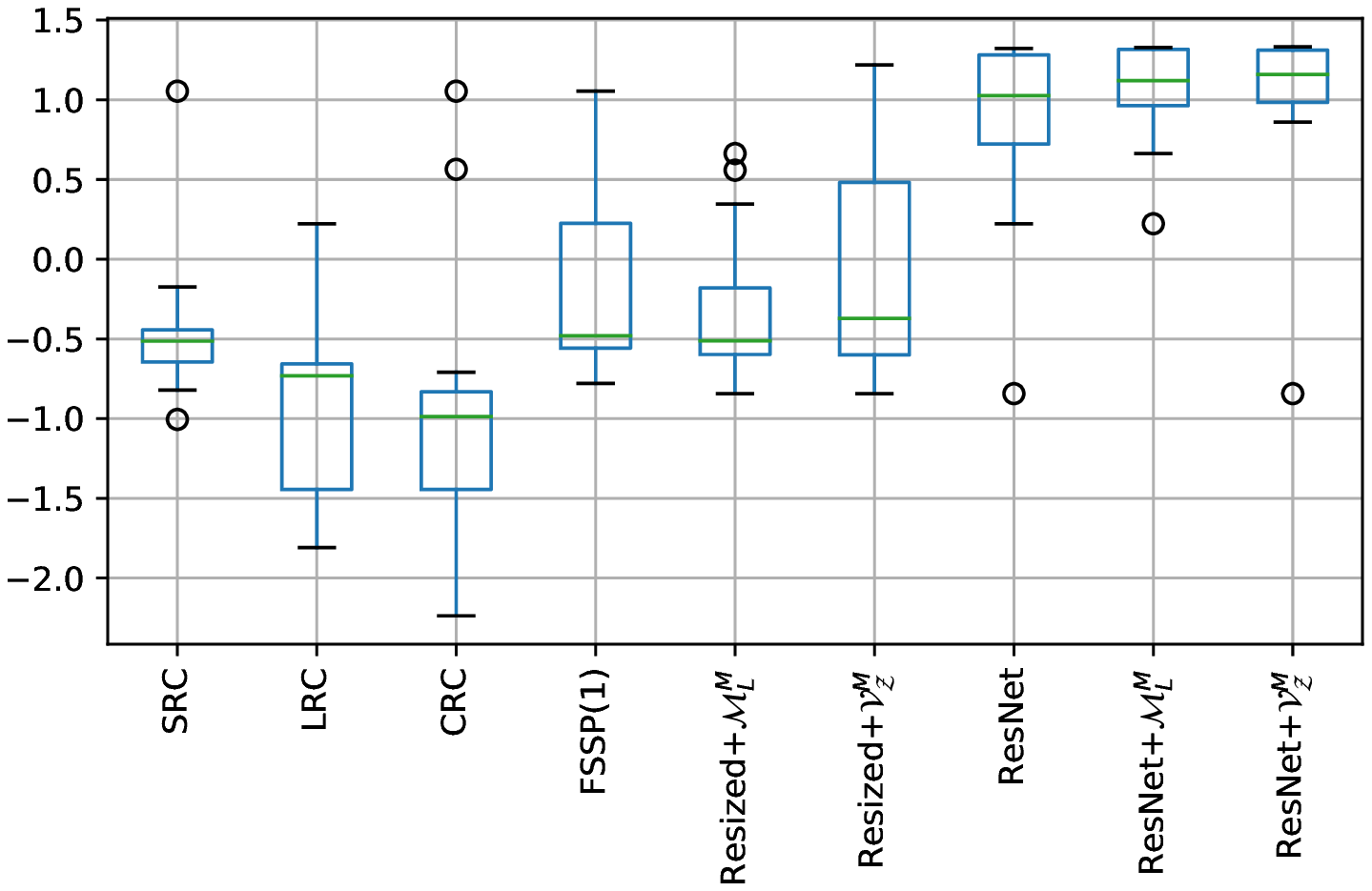} \\
b) Hasse diagram of Wilcoxson test.\\
\includegraphics[width=0.4\columnwidth]{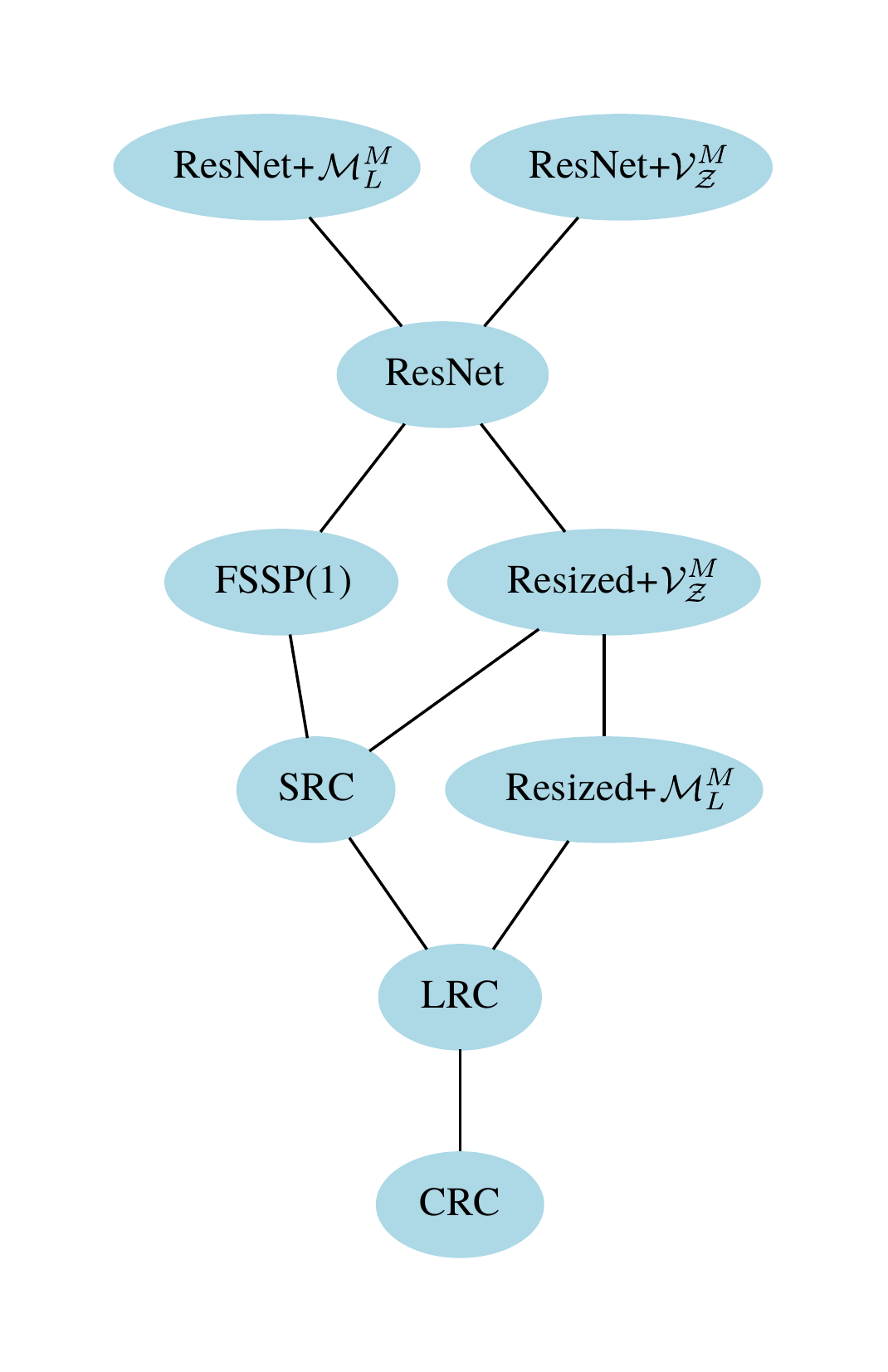}
\end{tabular}
 \caption{Box-plot and Hasse diagram of Wilcoxson signed-ranks test for the face recognition task.} \label{fig:boxplotAll}
\end{figure}

For the GT and the AT$\&$T face databases, we followed the ``first $N$'' scheme adopted by Feng et al. \cite{feng17}. Here, the first $N$ face images of each person are used as the training set. The remaining face images of each individual are used for test. The number $N$ varied according to the computational experiments described in \cite{feng17}. As to the AR face image database, we also followed the same evaluation protocol described in \cite{feng17}: Three facial expressions are used for train (e.g. normal, smile, and anger) while the remaining is used for test (e.g. scream). Tables \ref{tab:GT}, \ref{tab:ORL}, and \ref{tab:AR} list the recognition rates (RRs) yielded by the classifiers. These tables also provide the average recognition rate (ARR) for a given scenario. For a visual interpretation of the overall performance of the classifiers, Figure \ref{fig:boxplotAll}a) shows the box-plot comprising the normalized recognition rates listed on Tables \ref{tab:GT}, \ref{tab:ORL}, and \ref{tab:AR}. The normalized recognition rates are obtained by subtracting and dividing the values on Tables \ref{tab:GT}, \ref{tab:ORL}, and \ref{tab:AR} respectively by column-wise mean and standard deviation. Furthermore, Figure \ref{fig:boxplotAll}b) shows the Hasse diagram obtained from the outcome of the Wilcoxson signed-ranks test comparing any two classifiers with confidence level at 95\% using all the recognition rates listed on Tables \ref{tab:GT}, \ref{tab:ORL}, and \ref{tab:AR} \cite{burda13,demsar06,weise15}. Precisely, two classifiers are connected by an edge in this figure if the hypothesis test rejected the null hypothesis that the two classifiers perform equally well against the alternative hypothesis that the recognition rates of the classifier in the top is significantly larger than the recognition rates of the classifier on the bottom of the edge. In other words, the method on the top outperformed the method at the bottom of an edge. Also, we refrained to include the edges that can be derived from transitivity. For example, from Figure \ref{fig:boxplotAll}b) we have that $\mbox{ResNet}+\mathcal{V}_\mathcal{Z}^M$ is above ResNet and ResNet is above FSSP(1). Thus, we deduce that the $\mbox{ResNet}+\mathcal{V}_\mathcal{Z}^M$-based classifier outperformed the FSSP(1) in these experiments. 

Concluding, Figure \ref{fig:boxplotAll} shows that the ResNet$+\mathcal{V}_\mathcal{Z}^M$ and ResNet$+\mathcal{M}_L^M$-based classifiers outperformed all other classifiers, including the ResNet and FSSP(1) models, for recognition of uncorrupted face images. Recall, however, that the Zadeh's PAFMM $\mathcal{V}_\mathcal{Z}^M$ is computationally cheaper than AFMM $\mathcal{M}_L^M$. Let us now evaluate the performance of the classifiers in the presence of noise.


\begin{table}[t]
\begin{center}
\begin{tabular}{|l|l|l|l|l|l||l|} \hline 
Classifier  & $N=3$      & $N=4$      & $N= 5$ &  $N= 6$ & $N= 9$ &ARR \\ \hline
SRC         &  0.5367& 0.5836 &0.6240 &	0.7133 &  0.7867 		&0.6489 \\ \hline
LRC         & 0.5183 &0.5636  &0.5980&	0.6822&	0.7833 &0.6291\\ \hline
CRC         & 0.4683&0.5018	&0.5420 &	0.6200& 	0.7200		&0.5704  \\ \hline
FSSP(1)     & {0.5600}& {0.6000}  & {0.6300}&0.7044 &0.7800&  {0.6549}\\ \hline \hline
Resized+$\mathcal{M}_L^M$  &0.5567 & 0.5709  & 0.6000  & 0.7089 	&0.7900    & 0.6453 	\\ \hline
Resized+$\mathcal{V}_\mathcal{Z}^M$  &{0.5600}&0.5782&0.5900& 0.7222	& 0.8033& 0.6507	\\ \hline \hline 
ResNet  & \textbf{0.9533} & 0.9564 & 0.9560 & 0.9511 & {0.9600} & 0.9554  \\ \hline 
ResNet+$\mathcal{M}_L^M$  & \textbf{0.9533} & 0.9564 & 0.9580 & \textbf{0.9600} & \textbf{0.9633} & \textbf{0.9582}  \\ \hline
ResNet + $\mathcal{V}_\mathcal{Z}^M$ & \textbf{0.9533} & \textbf{0.9600} & \textbf{0.9600} & {0.9578} & {0.9600}  & \textbf{0.9582} \\ \hline \hline 
\end{tabular}
\caption{RRs and ARRs of the classifiers on GT face database with ``FIRST N'' scheme.} \label{tab:GT}
\end{center}
\end{table}

\begin{table}
\begin{center}
\begin{tabular}{|l|l|l|l|l|l||l|} \hline 
Classifier  &$N=3$ & $N=4$& $N= 5$& $N= 6$  &  $N= 7$&    ARR \\ \hline
SRC        & 0.8714 & 0.9167& 0.9300& 0.9500&	0.9583   & 0.9253  \\ \hline
LRC         & 0.8250 &0.8583	&0.9100& 0.9625&0.9583	 &  0.9028  \\ \hline
CRC         &0.8643 &0.9000 	&0.9100& 0.9187& 0.9250  & 0.9036   \\ \hline
FSSP(1)     & {0.9107} & {0.9417} & {0.9500}&0.9437& 0.9500& 0.9392 \\ \hline \hline
Resized+$\mathcal{M}_L^M$ & 0.8929  & 0.9042  & 0.9300  &  0.9688	&0.9667& 0.9325	\\ \hline
Resized+$\mathcal{V}_\mathcal{Z}^M$  &  0.8929	& 0.9167& {0.9500}& \textbf{0.9812}  &\textbf{0.9750 }  &  {0.9432} \\ \hline \hline 
ResNet & \textbf{0.9500} & 0.9625 & 0.9550 & 0.9625 & 0.9667 & 0.9593 \\ \hline 
ResNet+$\mathcal{M}_L^M$ & \textbf{0.9500} & \textbf{0.9708} & 0.9600 & 0.9625 & \textbf{0.9750} & 0.9637	\\ \hline
ResNet + $\mathcal{V}_\mathcal{Z}^M$ & \textbf{0.9500} & \textbf{0.9708} & \textbf{0.9650} & 0.9750 & \textbf{0.9750} & \textbf{0.9672} \\ \hline \hline
\end{tabular}
\caption{RRs and ARRs of the classifiers on  AT$\&$T face database with ``FIRST N'' scheme.} \label{tab:ORL}
\end{center}
\end{table}


\begin{table}
\begin{center}
\begin{tabular}{|l|l|l|l||l|} \hline 
Classifier  &Smile     & Anger    & Scream   & ARR \\ \hline
SRC        &	\textbf{1.000 }& 0.9800 	& 0.7900	& 0.9233\\ \hline
LRC         &0.9950	 &0.9700	&0.7650 &0.9100 \\ \hline
CRC         &\textbf{1.0000}	&{0.9950}	&0.7550	&0.9167\\ \hline
FSSP(1)     &\textbf{1.0000}	&0.9900	&0.8600	&0.9500\\ \hline \hline
Resized+$\mathcal{M}_L^M$ &  0.9950 & 0.9850  &0.9200   & {0.9667 }	 	\\ \hline
Resized+$\mathcal{V}_\mathcal{Z}^M$  &0.9950	&0.9800	&{0.9250}	& {0.9667} \\ \hline \hline 
ResNet & 0.9950 & \textbf{1.0000} & 0.9350 & 0.9767 \\ \hline
ResNet+$\mathcal{M}_L^M$ &  \textbf{1.0000} & \textbf{1.0000} & 0.9200 & 0.9733	 	\\ \hline
ResNet + $\mathcal{V}_\mathcal{Z}^M$ & 0.9950 & \textbf{1.0000} & \textbf{0.9450} & \textbf{0.9800} \\ \hline \hline
\end{tabular}
\caption{RRs and ARRs of the classifiers on AR face database with expressions.} \label{tab:AR}
\end{center}
\end{table}

\subsection{Face recognition in presence of noise}

\begin{figure}[t]
\begin{center}
\begin{tabular}{cccc}
\includegraphics[width=1.8cm, height = 2.2cm]{7} & \includegraphics[width=1.8cm, height = 2.2cm]{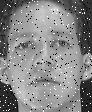}& \includegraphics[width=1.8cm, height = 2.2cm]{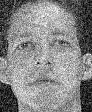}& \includegraphics[width=1.8cm, height = 2.2cm]{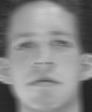}
\end{tabular}
\caption{Original images from AT$\&$T face database and versions corrupted respectively by salt and pepper noise, Gaussian noise, and horizontal motion (blurred image).}\label{fig:ATTnoise}
\end{center}
\end{figure}

In many practical situations, captured images are susceptible to different levels of noises and blurring effects. According to Gonzalez and  Woods \cite{gonzalez02}, the principal sources of noise in digital images arise during image acquisition and/or transmissions. The performance of imaging sensors is affected by a variety of factors such as environmental conditions and by the quality of the sensing elements themselves. For instance, Gaussian noise arises in an image due to factors as electronic circuit noise and sensor noise. On the other hand, salt and pepper noise is caused by transmission errors. Furthermore, a blurred image may arise either out of focus or relative motion between the camera and objects in the scene. Figure \ref{fig:ATTnoise} displays an undistorted and corrupted versions of an image from the AT$\&$T face database. The noise images have been obtained by introducing salt and pepper noise with probability $\rho=0.05$, Gaussian noise with mean 0 and variance $\sigma^2=0.01$, and by a horizontal motion of 9 pixels (blurred images).  

\begin{figure}[t]
\begin{center}
\subfigure{\includegraphics[width=0.75\linewidth]{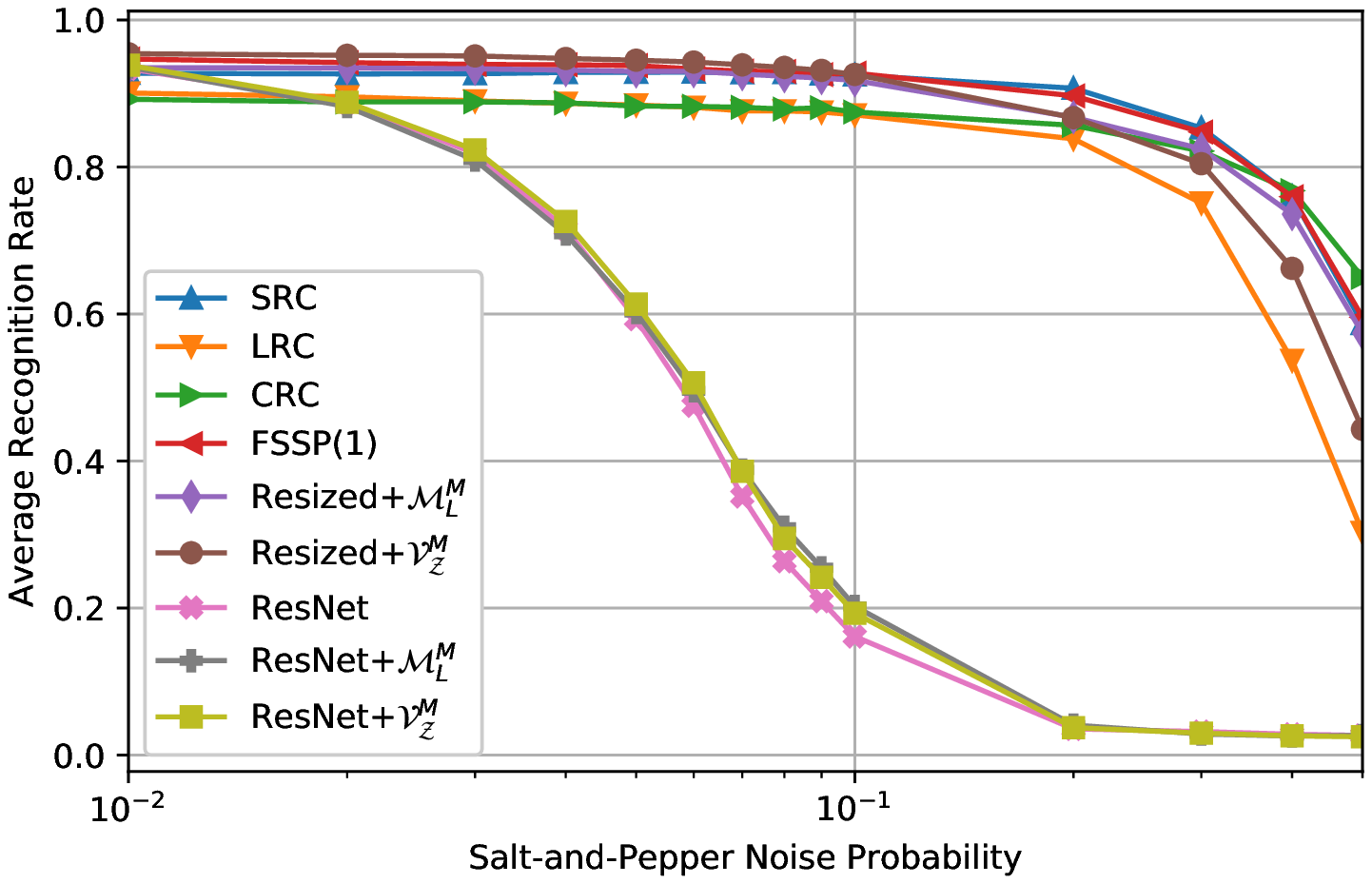}}
\subfigure{\includegraphics[width=0.75\linewidth]{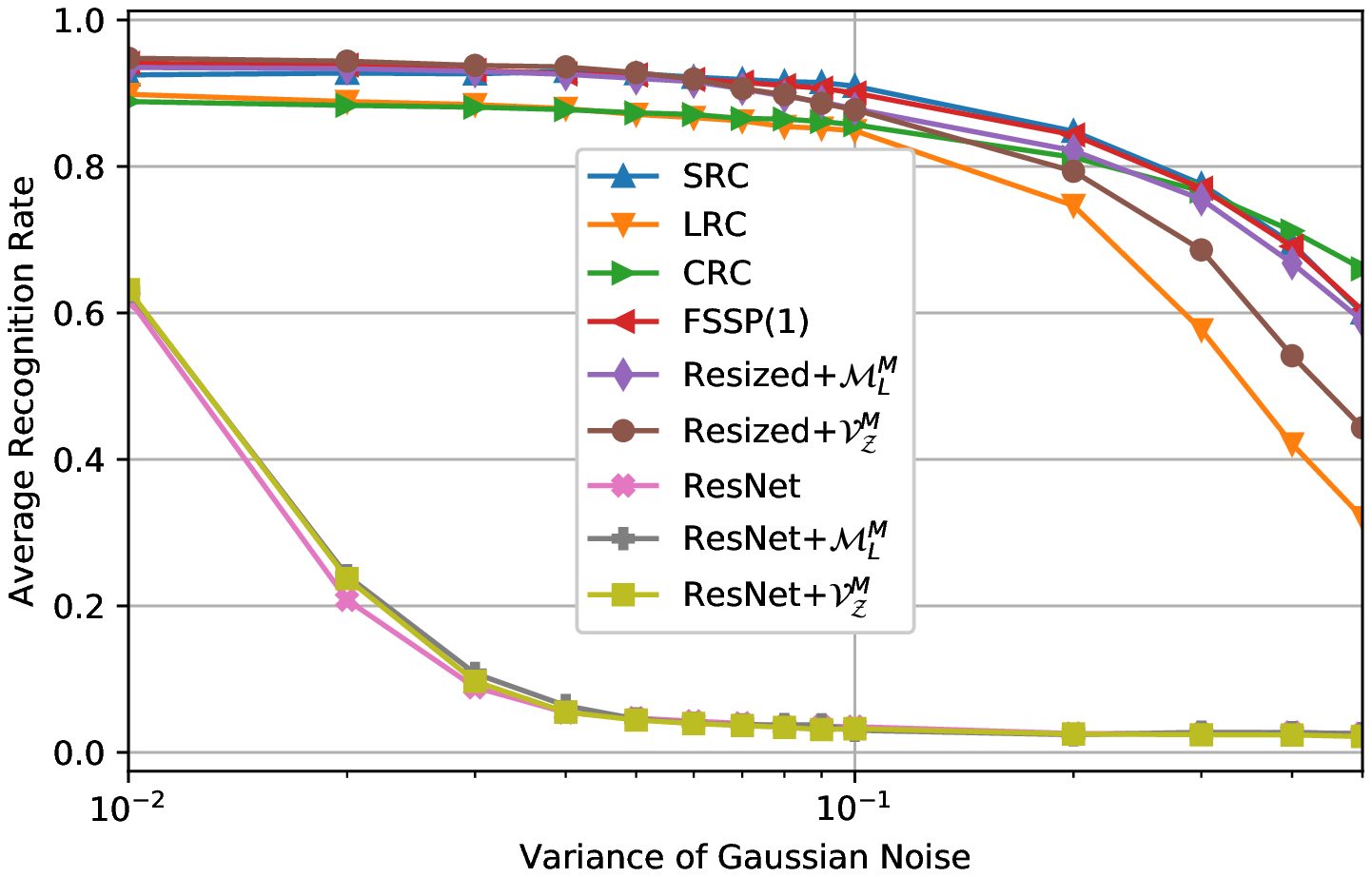}}
\subfigure{\includegraphics[width=0.75\linewidth]{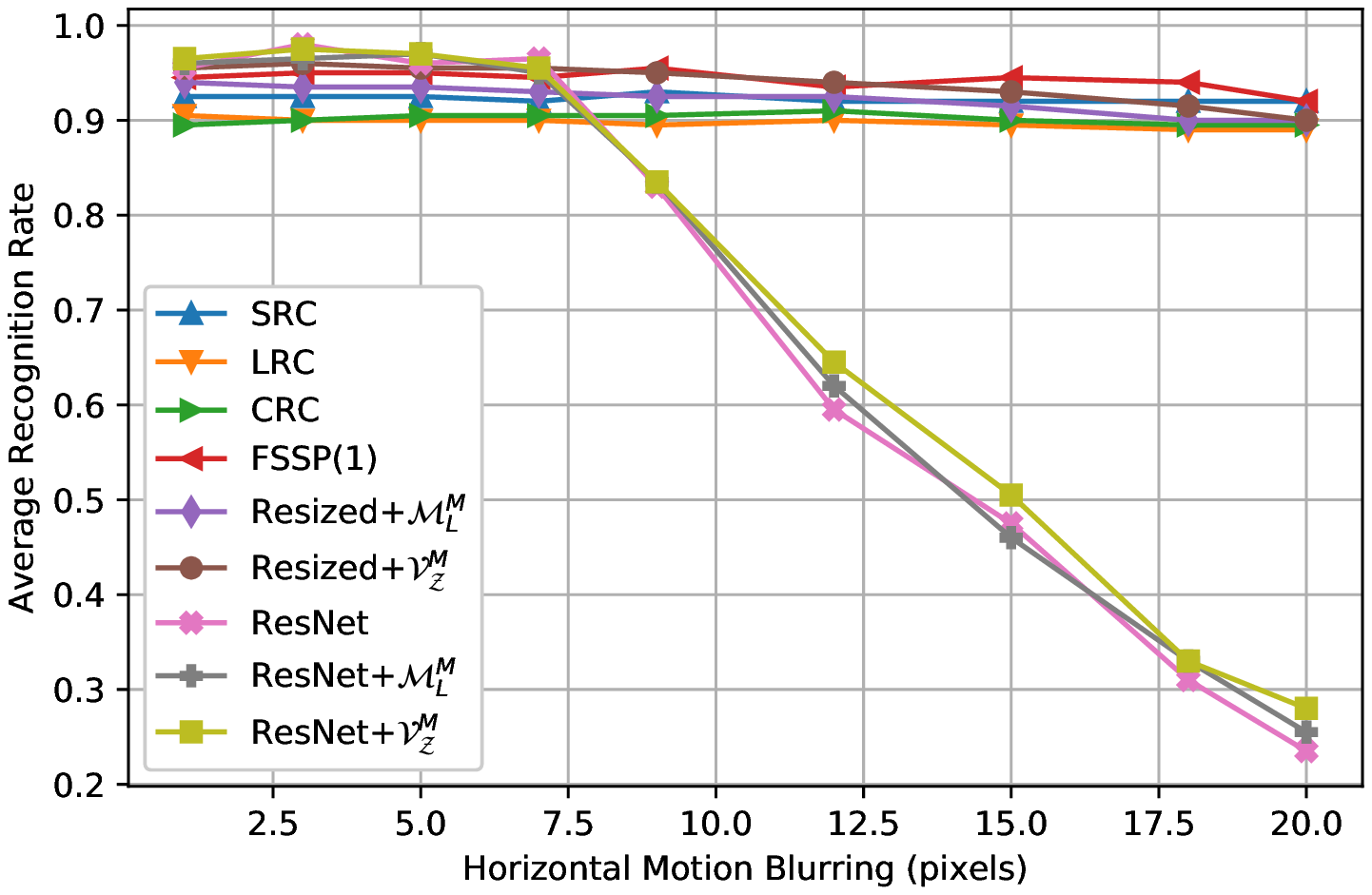}} 
\caption{Average recognition rate (ARR) versus noise intensity or horizontal motion.}\label{fig:Recognition}
\end{center}
\end{figure}

\begin{figure}[t]
\begin{center}
  \parbox[t]{0.3\columnwidth}{\centering a) Salt and pepper noise \includegraphics[width=0.33\columnwidth]{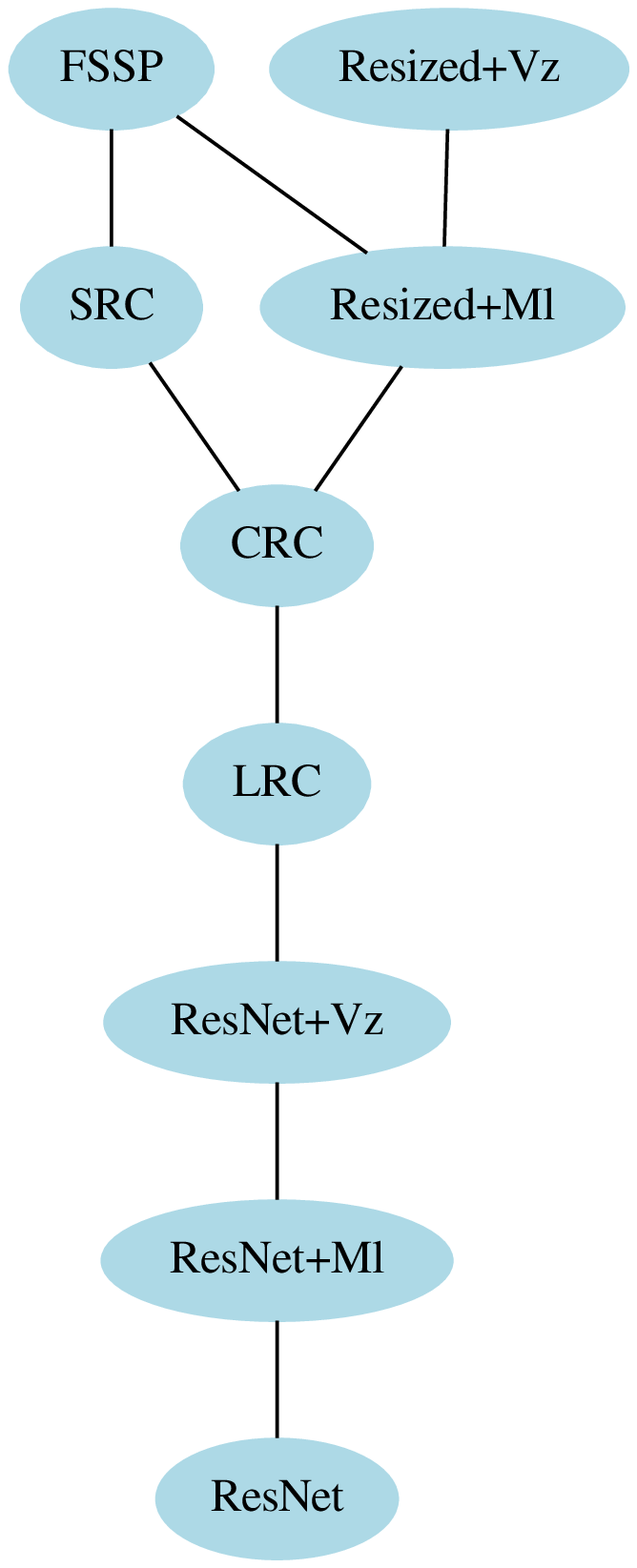}} 
  \parbox[t]{0.3\columnwidth}{\centering b) Gaussian noise \includegraphics[width=0.33\columnwidth]{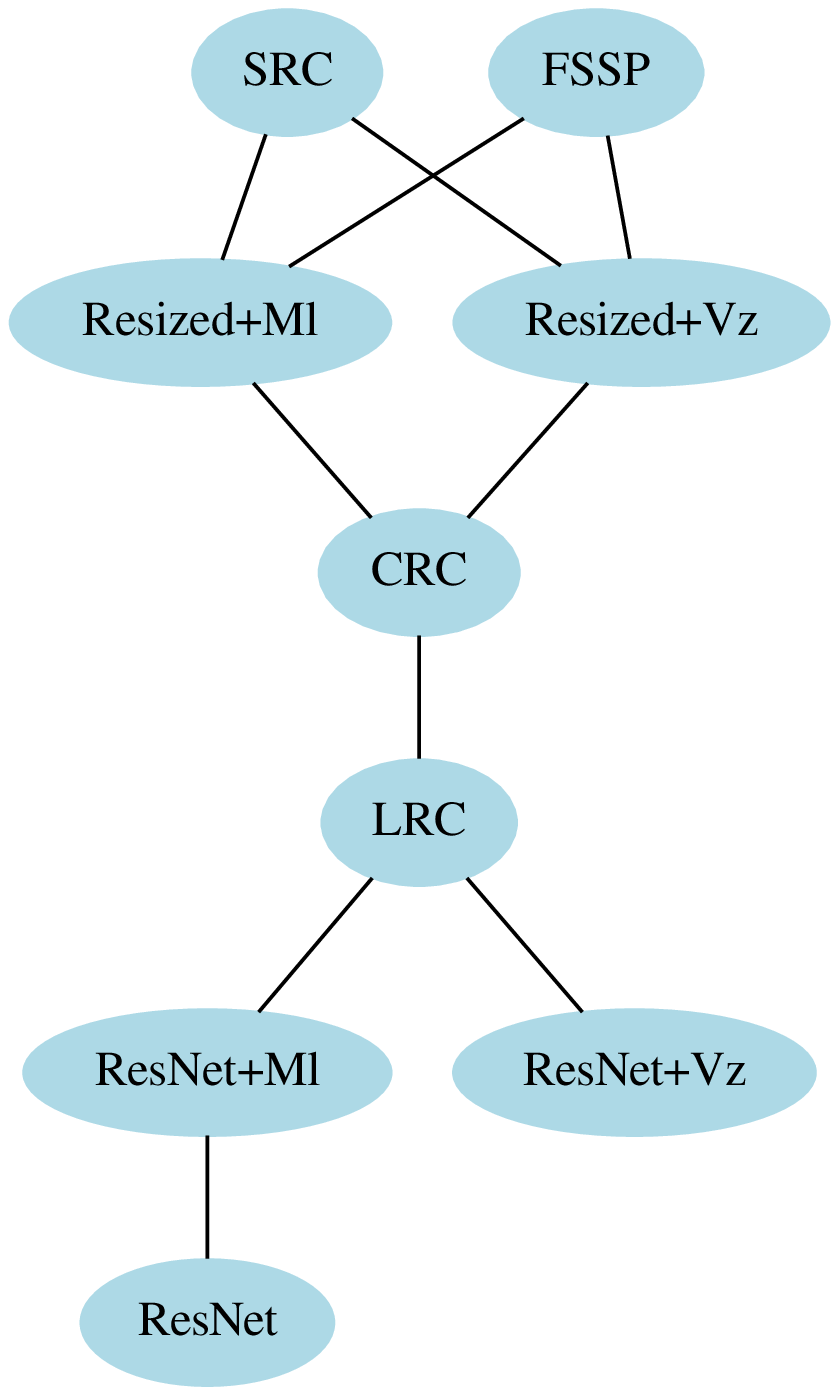}} 
  \parbox[t]{0.3\columnwidth}{\centering c) Blurred images \includegraphics[width=0.33\columnwidth]{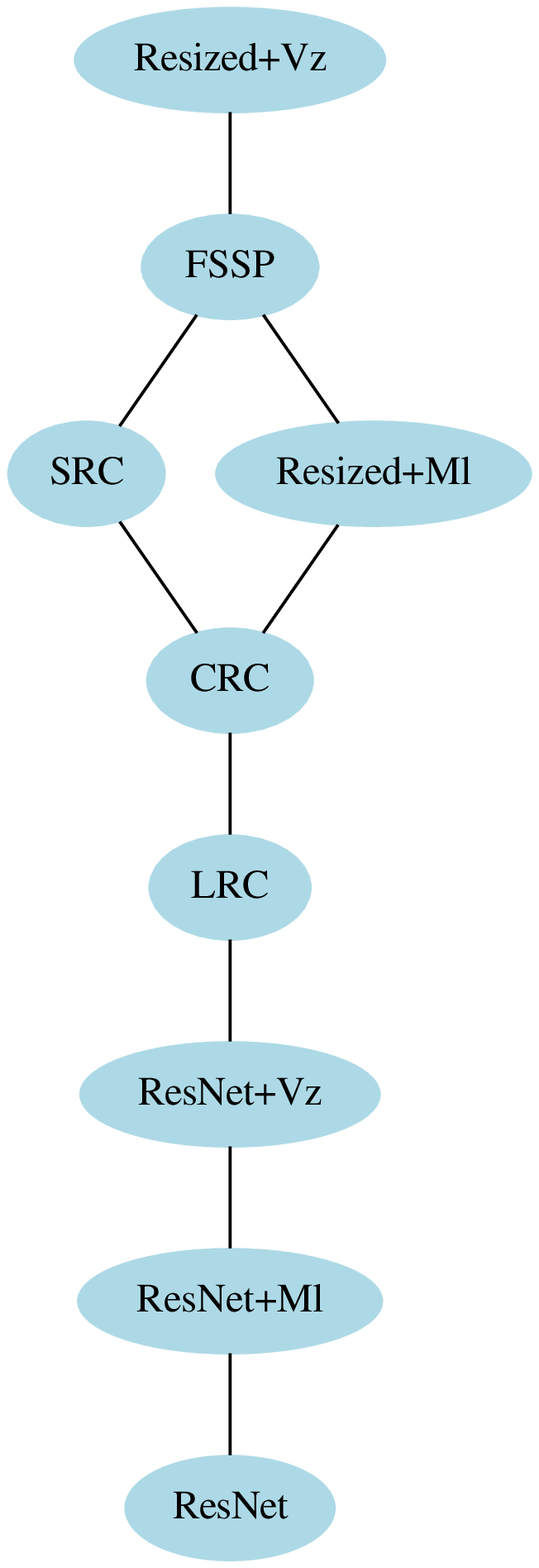}}
 \end{center}
 \caption{Hasse diagram of Wilcoxson signed-ranks test for the recognition task from corrupted input images.} \label{fig:HasseNoise}
\end{figure}

In order to simulate real-world conditions, we evaluated the performance of the classifiers when the training images are not distorted but test images are corrupted by some kind of noise. Precisely, test images have been corrupted by the following kind of noise:
\begin{enumerate}
  \item Salt and pepper noise with probability $\rho \in [0, 0.5]$; 
	\item  Gaussian noise with mean 0 and varience $\sigma^2 \in [0, 0.5]$;	
	\item Horizontal motion whose number of pixels varied from 1 to 20.
\end{enumerate}
In each scenario, the first 5 images of each individual of the AT$\&$T face database have used as training set and the remaining images, corrupted by some kind of noise, have been used for test. Figure \ref{fig:Recognition} shows the  average recognition rates (ARRs) produced by the eight classifiers in 30 experiments for each intensity of noise. Furthermore, Figure \ref{fig:HasseNoise} shows the Hasse diagram of the outcome of the Wilcoxson signed-ranks test comparing any two classifiers with confidence level at 95\%.  In contrast to the previous experiment with undistorted face images, the classifiers ResNet, ResNet+$\mathcal{V}_\mathcal{Z}^M$, and ResNet+$\mathcal{M}_L^M$ exhibited the worst recognition rates for corrupted input images. Moreover, we conclude the following from Figures \ref{fig:Recognition} and \ref{fig:HasseNoise}:
\begin{itemize}
 \item FSSP(1) and Resize+$\mathcal{V}_\mathcal{Z}^M$ are in general the best classifiers for the recognition of images corrupted by salt and pepper noise. 
 \item FSSP(1) and SRC yielded the larges recognition rates in the presence of Gaussian noise.
 \item Resized+$\mathcal{V}_\mathcal{Z}^M$ outperformed all the others classifiers for the recognition of blurred input images.
\end{itemize}
In general, Resized+$\mathcal{V}_\mathcal{Z}^M$ and FSSP(1) revealed to be the most robust classifiers for corrupted input images.

\subsection{Computational Complexity}

Let us conclude this section by analyzing the  computational complexity of the classifiers considered in this section. To this end, let $c$ denote the number of individuals, $k$ the number of training images per individual, and $n$ be either the number of pixels of the resized  face image or $n=128$ for vectors encoded by the ResNet. 

First of all, the following inequalities rank the computational complexity of the classifiers SRC, LRC, CRC and FSSP(1) \cite{feng17}:
\bb \mathcal{O}_{LRC}<\mathcal{O}_{CRC} < \mathcal{O}_{FSSP}<\mathcal{O}_{SRC}.\ee
Let us now compare the computational complexity of LRC and Resized+$\mathcal{V}_{\mathcal{Z}^M}$ classifiers.  On the one hand, the LRC is dominated by the solution of $c$ least square problems having $k$ unknowns and $n$ equations. Therefore, the computational complexity of the LRC method is
\bb \mathcal{O}_{LRC} = \mathcal{O}(cnk^2).\ee
On the other hand, due to the noise masking strategy, the complexity of the  Resized+$\mathcal{V}_\mathcal{Z}^M$ classifier is  
\bb \mathcal{O}_{\mbox{\scriptsize Resized}+\mathcal{V}_\mathcal{Z}^M}=\mathcal{O}(cnk).\ee
Thus, the Resized+$\mathcal{V}_\mathcal{Z}^M$ classifier is computationally cheaper than the LRC. According to Table \ref{tab:Floating}, the Resized+$\mathcal{V}_\mathcal{Z}^M$ classifier is also computationally cheaper then the Resized+$\mathcal{M}_L^M$ classifier. 
Therefore, the Resized-$\mathcal{V}_{\mathcal{Z}}^M$ classifier based is the cheapest among the classifiers based on resized versions of the face images.

As a pre-trained neural network, the computational effort to encode a face image into a vector of length $n=128$ is fixed. Discarding the encoding phase, the computational complexity of the ResNet classifier depends only on comparisons between the encoded input and the encoded trained vectors, which results $\mathcal{O}_{\mbox{\scriptsize ResNet}}= \mathcal{O}(ckn)$. In a similar fashion, we have $\mathcal{O}_{\mbox{\scriptsize Resized}+\mathcal{V}_\mathcal{Z}^M}=\mathcal{O}(cnk)$. In contrast, the computational effort of the ResNet+$\mathcal{M}_L^M$ is $\mathcal{O}(ck n^2)$, which is certainly more expensive than both ResNet and ResNet+$\mathcal{V}_{\mathcal{Z}}^M$.

Concluding, we believe that the $\mathcal{V}_\mathcal{Z}^M$-based classifiers are competitive with the state-of-the-art approaches because they exhibited a graceful balance between accuracy and computational cost.

\section{Concluding remarks}

In this paper, we investigated max-$C$ and min-$D$ projection autoassociative fuzzy memories (max-$C$ and min-$D$ PAFMMs). Briefly, PAFMMs are non-distributed versions of the well-know max-$C$ and min-$D$ autoassociative fuzzy morphological memories (max-$C$ and min-$D$ AFMMs) \cite{valle11nn,valle08fss}. Specifically, a PAFMM projects the input vector into either the family of all max-$C$ combinations or the family of min-$D$ combinations of the stored items. As a consequence, they present less spurious memories than their corresponding max-$C$ and min-$D$ AFMMs. Moreover, max-$C$ and min-$D$ PAFMMs are more robust to either dilative or erosive noise than the AFMMs. Besides, PAFMMs are computationally cheaper than AFMMs if the number of stored items $k$ is less that their length $n$. 

Apart from a detailed discussion on PAFMM models, in this paper we focused on the particular model referred to as Zadeh's max-$C$ PAFMM because it is obtained by considering Zadeh's fuzzy inclusion measure. Zadeh's max-$C$ and min-$D$ PAFMMs are the most robust PAFMMs with respect to either dilative or erosive noise. On the downside, they are extremely sensitive to mixed noise. In order to improve the noise tolerance of Zadeh's PAFMMs with respect to mixed noise, using a fuzzy similarity measure, we proposed a variation of the noise masking strategy of Urcid and Ritter \cite{urcid07LC}.

Finally, experimental results using three famous face databases confirmed the potential application of Zadeh's max-$C$ PAFMM for face recognition. Precisely, using Wilcoxson's signed-ranks test, we concluded that the ResNet+$\mathcal{V}_{\mathcal{Z}}^M$,  which is based on the ResNet encoding and the Zadeh's max-$C$ PAFMM classifier, outperformed important classifiers from literature including ResNet \cite{He16}, LRC \cite{naseem10}, CRC \cite{zhang11}, and FSSP(1) \cite{feng17} classifiers for the recognition of undistorted face images. Furthermore, the experiments revealed that the Resized+$\mathcal{V}_{\mathcal{Z}}^M$ classifier performs as well as the FSSP(1) \cite{feng17} method but requiring much less computational resources. 

In future, we intent to investigate further applications of max-$C$ and min-$D$ PAFMMs. In particular, we plan the study further the combinations of deep neural networks and these fuzzy associative memories. We also plan to generalize Zadeh's PAFMMs to more general complete lattices.

\section*{Acknowledgment}
This work was supported in part by CAPES -- Programa de Forma\c{c}\~ao Doutoral Docente, CNPq under grant no 305486/2014-4, and S\~ao Paulo Research Foundation (FAPESP) under grant no 2019/02278-2.

\appendix
\renewcommand*{\thesection}{Appendix \Alph{section}}
\section{Proofs of Theorems}

\begin{pot}

We are going to proof only the first part of Theorem \ref{theorem3}. The second part can be derived in a similar manner. Let $\vetz\in \mathcal{C}(\mathcal{A})$ be a max-$C$ combination of $\veta^1,\ldots,\veta^k$ and consider the set of indexes $\mathcal{N}=\{1,\ldots,n\}$ and $\mathcal{K}=\{1,\ldots,k\}$. Since the fuzzy implication $I$ and the fuzzy conjunction $C$ form an adjunction, we have:
\begin{align} 
\vetz \leq \vetx &\Longleftrightarrow \quad \bigvee_{\xi=1}^k C(\lambda_{\xi},a^{\xi}_j) \leq x_j, \ \forall j \in \mathcal{N} \\
&\Longleftrightarrow  \quad C(\lambda_{\xi},a^{\xi}_j) \leq x_j, \ \forall \xi \in \mathcal{K}, \forall j \in \mathcal{N} \\ 
& \Longleftrightarrow \quad \lambda_{\xi} \leq  I(a^{\xi}_j, x_j), \ \forall j \in \mathcal{N}, \ \forall \xi \in \mathcal{K}  \\
&\Longleftrightarrow \quad  \lambda_{\xi}\leq \bigwedge_{j=1}^n I(a^{\xi}_j, x_j), \ \forall \xi \in \mathcal{K}. \end{align}
Thus, the largest max-$C$ combination $\vetz=\bigvee_{\xi=1}^k C(\lambda_{\xi},\veta^{\xi})$ such that $\vetz \leq\vetx$ is obtained by considering $\lambda_{\xi}=\bigwedge_{j=1}^n I(a^{\xi}_j, x_j)$ for all $\xi \in \mathcal{K}$. 
\end{pot}

\begin{pot1}
 Let us only show \eqref{eq:dual}. The second part of the theorem is derived in a similar manner. First, recall that a strong negation is a decreasing operator. Thus, since the negation of the minimum is the maximum of the negations, we conclude from \eqref{eq:negation}, \eqref{forPontualS}, and \eqref{CdualD} that
\bbb
\mathcal{S}^{*}(\vetx) 
&=&\eta \Big(\mathcal{S}\big(\eta(\vetx) \big) \Big) 
=\eta \left( \bigwedge_{j=1}^n D(\theta_\xi, \veta^\xi) \right) \\
&=&\bigvee_{\xi=1}^k \eta\big(D(\theta_{\xi},\veta^{\xi})\big) 
=\bigvee_{\xi=1}^k C(\lambda_{\xi}^*,\mathbf{b}^{\xi}),\eee
where $\lambda_{\xi}^*=\eta(\theta_{\xi})$ satisfies the following identities
\bbb\lambda_{\xi}^* 
&= &\eta \left(\bigvee_{j=1}^n J \big(a_j^\xi,\eta(x_j) \big) \right)\\
&= &\bigwedge_{j=1}^n \eta \left(J \big(a_j^\xi,\eta(x_j) \big) \right)\\
&= &\bigwedge_{j=1}^n I \Big( \eta\big(a_j^\xi\big),x_j \Big)
= \bigwedge_{j=1}^n I \Big( b_j^\xi,x_j \Big).\eee
From \eqref{forPontual}, we have that $\mathcal{S}^*$ is the max-$C$ PAFMM designed for the storage of $\mathbf{b}^1,\ldots,\mathbf{b}^k$.
\end{pot1}

\begin{pot2} Consider the fundamental memory set $\mathcal{A}= \{\veta^1,\ldots,\veta^k\} \subset [0,1]^n$ and $\mathcal{K}=\{1,\ldots,k\}$.
If there exists an unique $\gamma\in \mathcal{K}$ such that $\veta^{\gamma} \leq \vetx$, then the index set, given by \eqref{eq:setJ}, is equal to 
  \bb\mathcal{I}=\{\xi: a_j^\xi \leq x_j, \forall j=1,\ldots,n\}= \left\{\gamma\right\}.\ee
Therefore,  \bb\mathcal{V}_{\mathcal{Z}}(\vetx)= \bigvee_{\xi \in \mathcal{I}_{\LL}}\veta^{\xi}=\veta^{\gamma}.\ee
Analogously, we can proof the second part of this theorem. 
\end{pot2}


\begin{thebibliography}{10}
\expandafter\ifx\csname url\endcsname\relax
  \def\url#1{\texttt{#1}}\fi
\expandafter\ifx\csname urlprefix\endcsname\relax\def\urlprefix{URL }\fi
\expandafter\ifx\csname href\endcsname\relax
  \def\href#1#2{#2} \def\path#1{#1}\fi

\bibitem{hassoun97}
M.~H. Hassoun, P.~B. Watta, {Associative Memory Networks}, in: E.~Fiesler,
  R.~Beale (Eds.), {Handbook of Neural Computation}, Oxford University Press,
  1997, pp. C1.3:1--C1.3:14.

\bibitem{kohonen89}
T.~Kohonen, {Self-organization and associative memory}, 3rd Edition,
  Springer-Verlag New York, Inc., New York, NY, USA, 1989.

\bibitem{hopfield85}
J.~Hopfield, D.~Tank, {Neural computation of decisions in optimization
  problems}, Biological Cybernetics 52 (1985) 141--152.

\bibitem{serpen08}
G.~Serpen, {Hopfield Network as Static Optimizer: Learning the Weights and
  Eliminating the Guesswork.}, Neural Processing Letters 27~(1) (2008) 1--15.
\newblock \href {http://dx.doi.org/10.1007/s11063-007-9055-8}
  {\path{doi:10.1007/s11063-007-9055-8}}.

\bibitem{valle11nn}
M.~E. Valle, P.~Sussner, {Storage and Recall Capabilities of Fuzzy
  Morphological Associative Memories with Adjunction-Based Learning}, Neural
  Networks 24~(1) (2011) 75--90.
\newblock \href {http://dx.doi.org/10.1016/j.neunet.2010.08.013}
  {\path{doi:10.1016/j.neunet.2010.08.013}}.

\bibitem{sussner18ins}
P.~Sussner, T.~Schuster, Interval-valued fuzzy morphological associative
  memories: Some theoretical aspects and applications, Information Sciences 438
  (2018) 127 -- 144.
\newblock \href {http://dx.doi.org/https://doi.org/10.1016/j.ins.2018.01.042}
  {\path{doi:https://doi.org/10.1016/j.ins.2018.01.042}}.

\bibitem{grana16}
M.~Grana, D.~Chyzhyk, Image understanding applications of lattice
  autoassociative memories, IEEE Transactions on Neural Networks and Learning
  Systems 27~(9) (2016) 1920--1932.
\newblock \href {http://dx.doi.org/10.1109/TNNLS.2015.2461451}
  {\path{doi:10.1109/TNNLS.2015.2461451}}.

\bibitem{sussner06fs}
P.~Sussner, M.~E. Valle, {Implicative Fuzzy Associative Memories}, IEEE
  Transactions on Fuzzy Systems 14~(6) (2006) 793--807.

\bibitem{sussner17msc}
P.~Sussner, M.~Ali, Image filters as reference functions for morphological
  associative memories in complete inf-semilattices, Mathware \& soft computing
  24 (2017) 53--62.

\bibitem{valle15}
M.~E. Valle, A.~C. Souza, On the recall capability of recurrent exponential
  fuzzy associative memories based on similarity measures, Mathware and Soft
  Computing Magazine 22 (2015) 33--39.

\bibitem{esmi12hais}
E.~L. Esmi, P.~Sussner, M.~E. Valle, F.~Sakuray, L.~Barros, {Fuzzy Associative
  Memories Based on Subsethood and Similarity Measures with Applications to
  Speaker Identification}, in: {Lecture Notes in Computer Science:
  International Conference on Hybrid Artificial Intelligence Systems (HAIS
  2012)}, Springer-Verlag Berlin Heidelberg, Berlin, Germany, 2012, pp.
  479--490.

\bibitem{esmi15fs}
E.~Esmi, P.~Sussner, H.~Bustince, J.~Fernandez, Theta-fuzzy associative
  memories (theta-fams), IEEE Transactions on Fuzzy Systems 23~(2) (2015)
  313--326.
\newblock \href {http://dx.doi.org/10.1109/TFUZZ.2014.2312131}
  {\path{doi:10.1109/TFUZZ.2014.2312131}}.

\bibitem{esmi16fss}
E.~Esmi, P.~Sussner, S.~Sandri, {Tunable equivalence fuzzy associative
  memories}, Fuzzy Sets and Systems 292~(Supplement C) (2016) 242 -- 260,
  special Issue in Honor of Francesc Esteva on the Occasion of his 70th
  Birthday.
\newblock \href {http://dx.doi.org/10.1016/j.fss.2015.04.004}
  {\path{doi:10.1016/j.fss.2015.04.004}}.

\bibitem{sussner06nn}
P.~Sussner, M.~E. Valle, {Grayscale Morphological Associative Memories}, IEEE
  Transactions on Neural Networks 17~(3) (2006) 559--570.

\bibitem{valle16gcsr}
M.~E. Valle, A.~C. Souza, Pattern classification using generalized recurrent
  exponential fuzzy associative memories, in: A.~G.~H. George A.~Papakostas,
  V.~G. Kaburlasos (Eds.), Handbook of Fuzzy Sets ComparisonHandbook of Fuzzy
  Sets Comparison Theory, Algorithms and Applications Theory, Algorithms and
  Applications, Vol.~6, Science Gate Publishing, 2016, Ch.~4, pp. 79--102.
\newblock \href {http://dx.doi.org/10.15579/gcsr.vol6.ch4}
  {\path{doi:10.15579/gcsr.vol6.ch4}}.

\bibitem{zhang04}
B.-L. Zhang, H.~Zhang, S.~S. Ge, {Face Recognition by Applying Wavelet Subband
  Representation and Kernel Associative Memory}, IEEE Transactions on Neural
  Networks 15~(1) (2004) 166--177.

\bibitem{kosko92}
B.~Kosko, {Neural Networks and Fuzzy Systems: A Dynamical Systems Approach to
  Machine Intelligence}, Prentice Hall, Englewood Cliffs, NJ, 1992.

\bibitem{goguen67}
J.~A. Goguen, {L-fuzzy sets}, Journal of Mathematical Analysis and Applications
  18 (1967) 145--174.

\bibitem{heijmans94}
H.~Heijmans, {Morphological Image Operators}, Academic Press, New York, NY,
  1994.

\bibitem{valle08fss}
M.~E. Valle, P.~Sussner, {A General Framework for Fuzzy Morphological
  Associative Memories}, Fuzzy Sets and Systems 159~(7) (2008) 747--768.

\bibitem{sussner11ins}
P.~Sussner, E.~L. Esmi, {Morphological Perceptrons with Competitive Learning:
  Lattice-Theoretical Framework and Constructive Learning Algorithm},
  Information Sciences 181~(10) (2011) 1929--1950.
\newblock \href {http://dx.doi.org/10.1016/j.ins.2010.03.016}
  {\path{doi:10.1016/j.ins.2010.03.016}}.

\bibitem{junbo94}
F.~Junbo, J.~Fan, S.~Yan, {A learning rule for fuzzy associative memories}, in:
  {Proceedings of the IEEE International Joint Conference on Neural Networks},
  Vol.~7, 1994, pp. 4273--4277.

\bibitem{liu99}
P.~Liu, {The Fuzzy Associative Memory of Max-Min Fuzzy Neural Networks with
  Threshold}, Fuzzy Sets and Systems 107 (1999) 147--157.

\bibitem{belohlavek00a}
R.~B\v{e}lohl{\'a}vek, {Fuzzy logical bidirectional associative memory},
  Information Sciences 128~(1-2) (2000) 91--103.

\bibitem{ritter98}
G.~X. Ritter, P.~Sussner, J.~L.~D. {de Leon}, {Morphological Associative
  Memories}, IEEE Transactions on Neural Networks 9~(2) (1998) 281--293.

\bibitem{vajg15}
M.~Vajgl, I.~Perfilieva, Associative memory in combination with the f-transform
  based image reduction, in: Fuzzy Systems (FUZZ-IEEE), 2015 IEEE International
  Conference on, 2015, pp. 1--6.
\newblock \href {http://dx.doi.org/10.1109/FUZZ-IEEE.2015.7338032}
  {\path{doi:10.1109/FUZZ-IEEE.2015.7338032}}.

\bibitem{valle10ins}
M.~E. Valle, {Permutation-Based Finite Implicative Fuzzy Associative Memories},
  Information Sciences 180~(21) (2010) 4136--4152.
\newblock \href {http://dx.doi.org/10.1016/j.ins.2010.07.003}
  {\path{doi:10.1016/j.ins.2010.07.003}}.

\bibitem{bui15}
T.~D. Bui, T.~H. Nong, T.~K. Dang, Improving learning rule for fuzzy
  associative memory with combination of content and association,
  Neurocomputing 149, Part A~(0) (2015) 59 -- 64, advances in neural networks
  Selected papers from the Tenth International Symposium on Neural Networks
  (ISNN 2013) Advances in Extreme Learning Machines Selected articles from the
  International Symposium on Extreme Learning Machines (ELM 2013).
\newblock \href {http://dx.doi.org/10.1016/j.neucom.2014.01.063}
  {\path{doi:10.1016/j.neucom.2014.01.063}}.

\bibitem{perfilieva15IFSA}
I.~Perfilieva, M.~Vajgl, Autoassociative fuzzy implicative memory on the
  platform of fuzzy preorder, in: Conference of the International Fuzzy Systems
  Association and the European Society for Fuzzy Logic and Technology
  (IFSA-EUSFLAT-15), Gij{\'{o}}n, Spain., June 30, 2015, pp. 1598--1603.

\bibitem{perfilieva16}
I.~Perfilieva, M.~Vajgl, Data retrieval and noise reduction by fuzzy
  associative memories, in: Proceedings of the Thirteenth International
  Conference on Concept Lattices and Their Applications, Moscow, Russia, July
  18-22, 2016, pp. 313--324.

\bibitem{vajgl17}
M.~Vajgl, Reduced ifam weight matrix representation using sparse matrices, in:
  Proceedings of: EUSFLAT- 2017 – The 10th Conference of the European Society
  for Fuzzy Logic and Technology and IWIFSGN’2017 – The Sixteenth
  International Workshop on Intuitionistic Fuzzy Sets and Generalized Nets,
  September 13-15, 2017, Warsaw, Poland, Volume 3, 2017, pp. 463--449.
\newblock \href {http://dx.doi.org/10.1007/978-3-319-66827-7 42}
  {\path{doi:10.1007/978-3-319-66827-7 42}}.

\bibitem{valle10nc}
M.~E. Valle, {Sparsely connected autoassociative fuzzy implicative memories and
  their application for the reconstruction of large gray-scale images},
  Neurocomputing 74~(1-3) (2010) 343--353.
\newblock \href {http://dx.doi.org/10.1016/j.neucom.2010.03.017}
  {\path{doi:10.1016/j.neucom.2010.03.017}}.

\bibitem{valle13prl}
M.~E. Valle, P.~Sussner, {Quantale-based autoassociative memories with an
  application to the storage of color images}, Pattern Recognition Letters
  34~(14) (2013) 1589--1601.

\bibitem{li17asc}
L.~Li, W.~Pedrycz, Z.~Li, Development of associative memories with transformed
  data, Applied Soft Computing (2017) --\href
  {http://dx.doi.org/10.1016/j.asoc.2017.05.035}
  {\path{doi:10.1016/j.asoc.2017.05.035}}.

\bibitem{ikeda01}
N.~Ikeda, P.~Watta, M.~Artiklar, M.~H. Hassoun, A two-level hamming network for
  high performance associative memory, Neural Networks 14~(9) (2001) 1189 --
  1200.
\newblock \href {http://dx.doi.org/10.1016/S0893-6080(01)00089-2}
  {\path{doi:10.1016/S0893-6080(01)00089-2}}.

\bibitem{souza18nafips}
A.~C. Souza, M.~E. Valle, Fuzzy kernel associative memories with application in
  classification, in: G.~A. Barreto, R.~Coelho (Eds.), Fuzzy Information
  Processing, Springer International Publishing, Cham, 2018, pp. 290--301.

\bibitem{souza18tema}
A.~C. Souza, M.~E. Valle, {Generalized Exponential Bidirectional Fuzzy
  Associative Memory with Fuzzy Cardinality-Based Similarity Measures Applied
  to Face Recognition}, Trends in Applied and Computational Mathematics 19~(2)
  (2018) 221 -- 233.

\bibitem{santos18nn}
A.~S. Santos, M.~E. Valle, Max-plus and min-plus projection autoassociative
  morphological memories and their compositions for pattern classification,
  Neural Networks 100 (2018) 84 -- 94.
\newblock \href {http://dx.doi.org/10.1016/j.neunet.2018.01.013}
  {\path{doi:10.1016/j.neunet.2018.01.013}}.

\bibitem{valle14wcciB}
M.~E. Valle, {An Introduction to Max-plus Projection Autoassociative
  Morphological Memory and Some of Its Variations}, in: {Proceedings of the
  IEEE International Conference on Fuzzy Systems 2014 {(FUZZ-IEEE 2014)}},
  Beijing, China, 2014, pp. 53--60.

\bibitem{santos16cbsf}
A.~S. Santos, M.~E. Valle, {Uma introdu\c c\~ao \`as mem\'orias
  autoassociativas fuzzy de proje\c c\~oes max-C}, in: { Recentes Avan\c cos em
  Sistemas Fuzzy. Sociedade Brasileira de Matem\'atica Aplicada e
  Computacional, Volume 1 }, S\~ao Carlos, Brasil, 2016, pp. 493--502, {ISBN:
  78-85-8215-079-5}.

\bibitem{santos17msc}
A.~S. Santos, M.~E. Valle, {The Class of Max-C Projection Autoassociative Fuzzy
  Memories}, Mathware and Soft Computing Magazine 24~(2) (2017) 63--73.

\bibitem{santos17cnmac}
A.~S. Santos, M.~E. Valle, {Some Theoretical Aspects of max-C and min-D
  Projection Fuzzy Autoassociative Memories }, in: {Proceeding Series of the
  Brazilian Society of Computational and Applied Mathematics 2017 ({CNMAC}
  2017)}, S{\~a}o Jos{\'e} dos Campos, Brazil, 2018, pp.~--.
\newblock \href {http://dx.doi.org/10.5540/03.2018.006.01.0436}
  {\path{doi:10.5540/03.2018.006.01.0436}}.

\bibitem{urcid07LC}
G.~Urcid, G.~X. Ritter, {Noise Masking for Pattern Recall Using a Single
  Lattice Matrix Associative Memory}, in: V.~Kaburlasos, G.~Ritter (Eds.),
  {Computational Intelligence Based on Lattice Theory}, Springer-Verlag,
  Heidelberg, Germany, 2007, Ch.~5, pp. 81--100.

\bibitem{santos17bracis}
A.~S. Santos, M.~E. Valle, {A Fast and Robust Max-C Projection Fuzzy
  Autoassociative Memory with an Application for Face Recognition }, in:
  {Proceedings of the Brazilian Conference on Intelligent Systems 2017
  ({BRACIS} 2017)}, Uberl{\^a}ndia, Brazil, 2017, pp. 306--311.
\newblock \href {http://dx.doi.org/10.1109/BRACIS.2017.57}
  {\path{doi:10.1109/BRACIS.2017.57}}.

\bibitem{barros17livro}
L.~C. {Barros}, R.~Bassanezi, W.~Lodwick, {First Course in Fuzzy Logic, Fuzzy
  Dynamical Systems, and Biomathematics,: Theory and Applications}, Vol. 347,
  Springer, 2017.

\bibitem{klir95}
G.~J. Klir, B.~Yuan, {Fuzzy Sets and Fuzzy Logic: Theory and Applications},
  Prentice Hall, Upper Saddle River, NY, 1995.

\bibitem{nguyen00}
H.~T. Nguyen, E.~A. Walker, {A First Course in Fuzzy Logic}, 2nd Edition,
  Chapman \& Hall/CRC, Boca Raton, 2000.

\bibitem{gomide07}
W.~Pedrycz, F.~Gomide, {Fuzzy Systems Engineering: Toward Human-Centric
  Computing}, Wiley-IEEE Press, New York, 2007.

\bibitem{debaets97a}
B.~{De Baets}, {Coimplicators, the forgotten connectives}, Trata Mountains
  Mathematical Publications 12 (1997) 229--240.

\bibitem{heijmans95}
H.~J. A.~M. Heijmans, {Mathematical Morphology: A Modern Approach in Image
  Processing Based on Algebra and Geometry}, SIAM Review 37~(1) (1995) 1--36.

\bibitem{deng02}
T.~Deng, H.~Heijmans, {Grey-scale morphology based on fuzzy logic}, Journal of
  Mathematical Imaging and Vision 16~(2) (2002) 155--171.

\bibitem{birkhoff93}
G.~Birkhoff, {Lattice Theory}, 3rd Edition, American Mathematical Society,
  Providence, 1993.

\bibitem{davey02}
B.~Davey, H.~Priestley, Introduction to Lattices and Order (2nd ed.), Cambridge
  University Press, 2002.

\bibitem{blyth72}
T.~Blyth, M.~Janowitz, {Residuation Theory}, Pergamon Press, Oxford, 1972.

\bibitem{bandler80}
W.~Bandler, L.~Kohout, {Fuzzy power sets and fuzzy implication operators},
  Fuzzy Sets and Systems 4~(1) (1980) 13--30.

\bibitem{sussner08gr}
P.~Sussner, M.~E. Valle, {Fuzzy Associative Memories and Their Relationship to
  Mathematical Morphology}, in: W.~Pedrycz, A.~Skowron, V.~Kreinovich (Eds.),
  {Handbook of Granular Computing}, John Wiley and Sons, Inc., New York, 2008,
  Ch.~33, pp. 733--754.

\bibitem{valle18wiley}
M.~E. Valle, P.~Sussner, A.~S. Santos, Morphological associative memories, in:
  {Webster, John G.} (Ed.), {Wiley Encyclopedia of Electrical and Electronics
  Engineering}, John Wiley \& Sons, Inc., 2018, pp.~--.
\newblock \href {http://dx.doi.org/10.1002/047134608X.W8363}
  {\path{doi:10.1002/047134608X.W8363}}.

\bibitem{couso13}
I.~Couso, L.~Garrido, L.~Sánchez, Similarity and dissimilarity measures
  between fuzzy sets: A formal relational study, Information Sciences 229
  (2013) 122 -- 141.
\newblock \href {http://dx.doi.org/https://doi.org/10.1016/j.ins.2012.11.012}
  {\path{doi:https://doi.org/10.1016/j.ins.2012.11.012}}.

\bibitem{debaets05}
B.~{De Baets}, H.~{De Meyer}, Transitivity-preserving fuzzification schemes for
  cardinality-based similarity measures, European Journal of Operational
  Research 160~(3) (2005) 726 -- 740, dOI:10.1016/j.ejor.2003.06.036.

\bibitem{baets09}
B.~{De Baets}, S.~Janssens, H.~{De Meyer}, On the transitivity of a parametric
  family of cardinality-based similarity measures, International Journal of
  Approximate Reasoning 50~(1) (2009) 104 -- 116, special Section on Recent
  advances in soft computing in image processing and Special Section on
  Selected papers from NAFIPS 2006.
\newblock \href {http://dx.doi.org/https://doi.org/10.1016/j.ijar.2008.03.006}
  {\path{doi:https://doi.org/10.1016/j.ijar.2008.03.006}}.

\bibitem{fan99a}
J.~Fan, W.~Xie, Some notes on similarity measure and proximity measure, Fuzzy
  Sets and Systems 101~(3) (1999) 403 -- 412.
\newblock \href
  {http://dx.doi.org/https://doi.org/10.1016/S0165-0114(97)00108-5}
  {\path{doi:https://doi.org/10.1016/S0165-0114(97)00108-5}}.

\bibitem{xuecheng92}
L.~Xuecheng, {Entropy, distance measure and similarity measure of fuzzy sets
  and their relations}, Fuzzy Sets and Systems 52~(3) (1992) 305--318.
\newblock \href {http://dx.doi.org/10.1016/0165-0114(92)90239-Z}
  {\path{doi:10.1016/0165-0114(92)90239-Z}}.

\bibitem{feng17}
Q.~Feng, C.~Yuan, J.~S. Pan, J.~F. Yang, Y.~T. Chou, Y.~Zhou, W.~Li,
  Superimposed sparse parameter classifiers for face recognition, IEEE
  Transactions on Cybernetics 47~(2) (2017) 378--390.
\newblock \href {http://dx.doi.org/10.1109/TCYB.2016.2516239}
  {\path{doi:10.1109/TCYB.2016.2516239}}.

\bibitem{He16}
K.~{He}, X.~{Zhang}, S.~{Ren}, J.~{Sun}, Deep residual learning for image
  recognition, in: 2016 IEEE Conference on Computer Vision and Pattern
  Recognition (CVPR), 2016, pp. 770--778.
\newblock \href {http://dx.doi.org/10.1109/CVPR.2016.90}
  {\path{doi:10.1109/CVPR.2016.90}}.

\bibitem{Dlib}
D.~E. King, Dlib-ml: A machine learning toolkit, Journal of Machine Learning
  Research 10 (2009) 1755--1758.

\bibitem{wright09}
J.~Wright, A.~Y. Yang, A.~Ganesh, S.~S. Sastry, Y.~Ma, Robust face recognition
  via sparse representation, IEEE Transactions on Pattern Analysis and Machine
  Intelligence 31~(2) (2009) 210--227.
\newblock \href {http://dx.doi.org/10.1109/TPAMI.2008.79}
  {\path{doi:10.1109/TPAMI.2008.79}}.

\bibitem{naseem10}
I.~Naseem, R.~Togneri, M.~Bennamoun, Linear regression for face recognition,
  IEEE Transactions on Pattern Analysis and Machine Intelligence 32~(11) (2010)
  2106--2112.
\newblock \href {http://dx.doi.org/10.1109/TPAMI.2010.128}
  {\path{doi:10.1109/TPAMI.2010.128}}.

\bibitem{zhang11}
L.~Zhang, M.~Yang, X.~Feng, Sparse representation or collaborative
  representation: Which helps face recognition?, in: 2011 International
  Conference on Computer Vision, 2011, pp. 471--478.
\newblock \href {http://dx.doi.org/10.1109/ICCV.2011.6126277}
  {\path{doi:10.1109/ICCV.2011.6126277}}.

\bibitem{GTdatabase}
A.~V. Nefian, {Georgia Tech face database}, {Avaliable
  at:http://www.anefian.com/research/face$\_$reco.htm} (2017).

\bibitem{ORLdatabase}
A.~L. Cambridge, {The AT{\&}T Database of Faces}, {Avaliable at:
  http://www.cl.cam.ac.uk/research/dtg/attarchive/facedatabase.html} (1994).

\bibitem{martinez98a}
A.~M. Martinez, R.~Benavente, {The AR Face Database}, Tech. Rep.~24, CVC (Jun.
  1998).

\bibitem{burda13}
M.~Burda, paircompviz: An {R} package for visualization of multiple pairwise
  comparison test results (2013).
\newblock \href {http://dx.doi.org/10.18129/B9.bioc.paircompviz}
  {\path{doi:10.18129/B9.bioc.paircompviz}}.

\bibitem{demsar06}
J.~Demšar, Statistical comparisons of classifiers over multiple data sets,
  Journal of Machine Learning Research 7 (2006) 1--30.

\bibitem{weise15}
T.~Weise, R.~Chiong, An alternative way of presenting statistical test results
  when evaluating the performance of stochastic approaches, Neurocomputing 147
  (2015) 235 -- 238.
\newblock \href {http://dx.doi.org/10.1016/j.neucom.2014.06.071}
  {\path{doi:10.1016/j.neucom.2014.06.071}}.

\bibitem{gonzalez02}
R.~C. Gonzalez, R.~E. Woods, {Digital Image Processing}, 2nd Edition,
  Prentice-Hall, Upper Saddle River, NJ, 2002.

\end{thebibliography}



%

%
%

\end{document}